\documentclass[preprint]{elsarticle}

\usepackage{hyperref}
\usepackage{stdlib}
\usepackage[section]{placeins}
\usepackage{changepage}

\journal{Artificial Intelligence Journal}

\begin{document}

\begin{frontmatter}

\title{SMT-based Robot Transition Repair}

\author{Jarrett Holtz,
Arjun Guha, {\normalfont and}
Joydeep Biswas}
\ead{\{jaholtz,arjun,joydeepb\}@cs.umass.edu}
\address{University of Massachusetts Amherst, CICS \\ 140 Governors Drive, Amherst, MA 01002, USA}




\begin{abstract}
State machines are a common model for robot behaviors that combine a number
of individual controllers with a transition function that switches between them. 
Transition
functions often rely on parameterized conditions to model preconditions
for the controllers, where
the correct values of the parameters depend on factors relating
to the environment or the specific robot. In the absence
of specific calibration procedures a roboticist
must painstakingly adjust the parameters through
a series of trial and error experiments. In this process, identifying
when the robot has taken an incorrect action, and what should be
done instead is straightforward, but finding the right parameter values
can be difficult.
Inspired by this idea we present an
alternative approach that we call, 
\emph{interactive SMT-based Robot Transition Repair} (\technique).
During exection we record an \textit{execution trace} 
of the transition function, and  we ask the roboticist to identify a
few instances where the robot has transitioned incorrectly, and what the 
correct transition should have been. 
A user supplies these \textit{corrections} based on
the type of error to repair. Either by directly identifying them in
the trace when a transition should occur, 
or by generating
corrections via forward simulation when a transition should not occur.
Using these corrections,
an automated analysis of the traces \textit{partially evaluates} the 
transition function for each correction. Simplifying the transition function
to a system of constraints that describe the correct behavior in terms
of the repairable parameters. 
This system of constraints is then formulated as a \maxsmt{} problem, 
where the solution is a minimal
adjustment to the parameters that satisfies the maximum number of constraints. 
In order to identify a repair
that accurately captures user intentions and generalizes to
novel scenarios, solutions are explored by iteratively adding constraints to the
\maxsmt{} problem to yield sets of alternative repairs.
We test with state machines from multiple domains
including robot soccer and autonomous driving, and we evaluate solver based repair
with respect to solver choice and optimization hyperparameters. 
Our results demonstrate
that \technique{} can repair a variety of states machines and error types
1) quickly, 2) with small numbers of corrections, while 3) not overcorrecting
state machines and harming generalized performance. Finally, we show that 
a state machine corrected with \technique{} can outperform an expert-tuned
state machine deployed in a real-world scenario.

%

\end{abstract}


\end{frontmatter}


\section{Introduction}
In robotics complex behaviors often use state machines to combine
feedback controllers as \textit{states} by using a transition function to
select the correct actions.
Even when the individual controllers performs as intended,
the transitions between states are often dependent on a set of parameters
that require tuning to maximize performance. It is often the case
that a single set of parameters will be optimal on one robot, or in one environment,
but fail when transferred to another. \figref{attacker-examples} shows an
example failure using two trajectories of a
robotic soccer player as it tries to kick a moving ball. A very small change
to its parameter values is the difference between success and failure.

No single good solution to this problem exists. For special cases, calibration
procedures can be used to adjust parameters automatically
(\eg{} \cite{holtz2017automatic}), but these are application specific. More
general optimization techniques are subject to local minimal because robot performance
is often non-convex with respect to parameter values, and exhaustive-search,
while applicable, is impractical even for relatively simple robots. Therefore,
manually adjusting parameters to iteratively improve performance is the primary
approached used by roboticists.

\begin{figure}
 \centering
 \begin{subfigure}[b]{0.49\linewidth}
 \centering
 \begin{tikzpicture}[scale=0.05]
  \tikzstyle{every node}+=[rounded corners,minimum width=0.9cm,minimum
  height=0.35cm,node distance=0.8cm and 1.7cm,draw=black,inner
  sep=0pt,font=\tiny\itshape]
  \tikzstyle{every path}+=[>=latex]
  \node[fill=blue!25] (start) {Start};
  \node (goto) [below of=start] {Go To};
  \coordinate[below of=goto] (belowgoto);
  \node (intercept) [right of=belowgoto] {Intercept};
  \node (catch) [left of=belowgoto] {Catch};
  \node (kick) [below of=belowgoto] {Kick};
  \node[fill=green!25] (end) [below of=kick] {End};
  \path[every loop/.style={looseness=5}]
    (start) [->] edge (goto)
    (goto) [<->] edge (kick)
    (goto) [<->] edge (intercept)
    (goto) [<->] edge (catch)
    (intercept) [<->] edge (kick)
    (catch) [<->] edge (kick)
    (catch) [<->] edge (intercept)
    (kick) [->] edge (end)
    (goto) [->] edge [loop left] (goto)
    (catch) [->] edge [loop left] (catch)
    (intercept) [->] edge [loop right] (intercept)
    (kick) [->] edge [loop left] (kick);
  \end{tikzpicture}
  \caption{Attacker state machine}
  \figlabel{example-model}
 \end{subfigure}
 \begin{subfigure}[b]{0.4\linewidth}
  \includegraphics[width=0.85\linewidth]{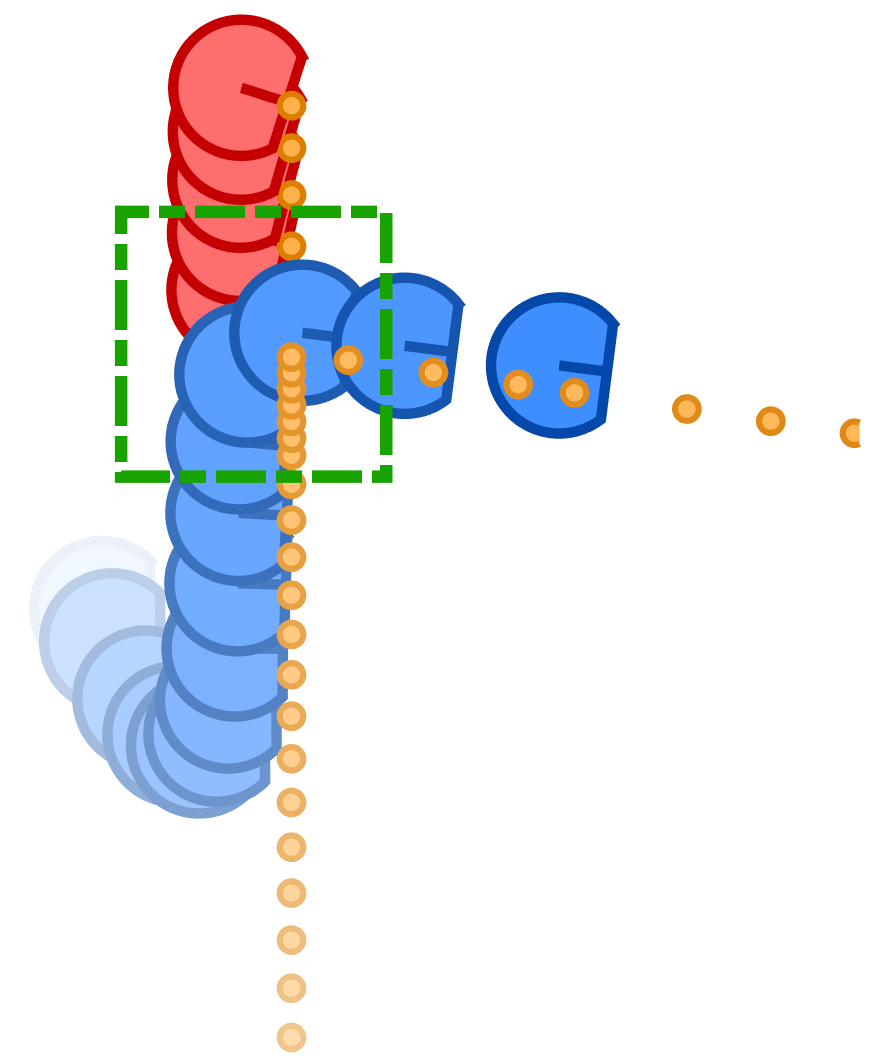}
  \caption{Execution traces}
  \figlabel{example-trajectories}
 \end{subfigure}
 \caption{A robot soccer attacker a) state machine, with b)
   successful (blue) and unsuccessful (red) traces to {\itshape Intercept}
   a ball (orange) and {\itshape Kick} at the goal. The green box isolates the error: the
  successful trace transitions to {\itshape Kick}, the unsuccessful trace remains
  in {\itshape Intercept}.}
  \figlabel{attacker-examples}
\end{figure}
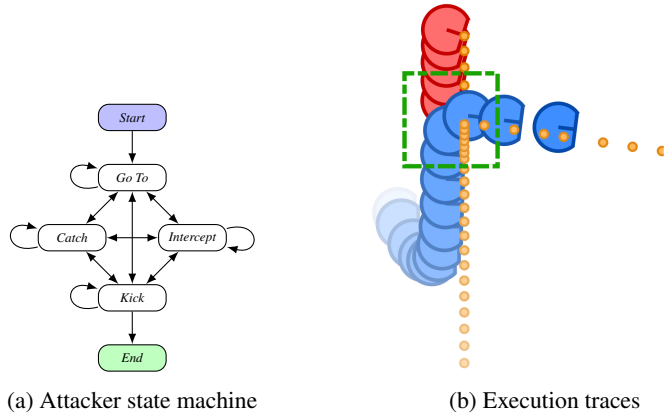

In manually tuning these values the first step a roboticist takes is
to identify what has gone wrong, and what the robot should have done instead.
In comparison to the tedious task of identifying the correct parameter
values, this is a simple task. When debugging a state machine for robot
control this requires a roboticist to identify when the robot is in an
incorrect state, and what the correct state should be. This results
in a \emph{partial specification} of the intended behavior: it specifies
only a subset of the correct behavior without any information about how
to realize that behavior. These partial specifications
are the first step in the standard iterative adjustment procedure for robot behaviors,
but actually realizing the desired behavior through iterative adjustment is much
more difficult.

In this article, we present \emph{Interactive SMT-based Robot Transition Repair} (\technique{})
a technique for treating these partial
specifications as user \emph{corrections}, and we show that these corrections alone
are adequate inputs for an automatic parameter repair procedure that automatically realizes
the desired behavior without additional user effort. This repair
procedure leverages
Satisfiable Modulo Theories (SMT) in order to model transition functions and
corrections in first-order logic. SMT is an expressive model of computation that is
commonly used in applications such as program verification, synthesis, and a model checking.
With the addition of optimizing SMT solvers \cite{bjorner2015nuz} SMT can be used to
complete models of computation while also optimizing objective functions. While
least squares optimization or mixed integer linear programming are more popular
optimization techniques in robotics, they cannot model problems
that contain conditionals
that SMT can. SMT based optimization uses techniques similar to MILP solvers
for optimization, but combines arithmetic optimization and SMT solving
to model a wider variety of problems, including those with conditionals. In some
cases these problems may reduce to MILP problems, but SMT solvers can find
solutions even in these edge cases.

In order to construct and solve the first-order logic problem
\technique{} first logs an execution trace of the transition function
that captures program structure, program state, and information about the world
relevant to the transition function.
After execution, the roboticist examines the trace and
provides one or more corrections. Based on the type of error a user can
provide either \textit{\imm{s}} that identify exact desired
behavior in the execution trace, or utilize \textit{\continue{s}} to fork
the world state and generate a sequence of corrections that delay a desired
transition (\secref{corrections}).
\technique{} then utilizes information from the execution trace, analysis
of the transitions function, and the set of user corrections to construct
a formula in first-order logic suitable for a solver (\secref{max-smt}).
When a user supplies
more than one correction they may not all be satisfiable, and so there may
be several possible models that partially satisfy user intent. In order
to explore these solutions \technique{} iteratively updates and solves a \maxsmt{}
problem that seeks to maximize the number of satisfied corrections while minimizing
the magnitude of the repairs to the parameters (\secref{solving}).
At each iteration the \maxsmt{}
problem is updated with additional constraints that guide the solver to novel
solutions such that a set of potential solutions is presented to the user.

To evaluate \technique{} we use it as a repair procedure for state machines in
a number of domains, and we evaluate properties of the \maxsmt{} formulation
and solvers used, as well as the overall performance of \technique{} both
with respect to success rate and solution time.
We demonstrate experimentally that \technique{}:
\begin{inparaenum}[1)]
\item is more computationally tractable than exhaustive parameter search;
\item is scalable and can generate meaningful repairs with different backend solvers;
\item can be applied to various state machines in different domains;
\item generalizes to unseen situations;
\item can provide robust repairs even in special cases where the trace and corrections
do not provide a single ideal repair; and
\item when applied
 to simple robot soccer attacker, can
  outperform a more complicated, expert-tuned attacker that won the
  lower bracket finals at \emph{RoboCup 2017}.
\end{inparaenum}

\section{Related Work}
\seclabel{related}

Configuration parameters are common in many software systems,
and represent a large source of error across disciplines
\cite{xu2015systems,weiss2017tortoise,cano2016automatic}.
In robotics, behaviors often rely on environment-dependent parameters for robust
and accurate execution.
Using models of the desired structure and properties of behaviors,
robot software can be designed at a higher level of abstraction
that allows for automatic parameter adaptation to hardware, software,
and architectural changes to achieve system objectives~\cite{brassModelBased}.
If a precise model of the dependency between parameters and
behaviors is available, it may be possible to design
a \emph{calibration procedure} that executes a specific sequence of actions and
to recover correct parameter
values~(\eg{}\cite{holtz2017automatic}). If a calibration procedure
cannot be designed, but the effect of parameters
is well-understood, it may be possible to optimize for the parameters using a
functional model~\cite{cano2016automatic}.
Model-based diagnosis can diagnose faulty parameters~\cite{reiter1987theory} if
the behavior of the robot in its environment can be formally defined.
Properties such as liveness,
timeliness, etc can be modeled for monitoring correctness as in RoboChart~\cite{miyazawa2017},
which uses finite state machines to verify behaviors.
Similar to our approach is \cite{taylor2016codiagnosing}, which uses human
input to identify error types in robot behavior and attempts to identify
predicates in the code that are relevant to the failure in order to localize
the faults.

In systems, approaches to handling
system configuration error fall into two major categories
designing systems to negate or minimize the impact of configuration errors,
or supplying user end tools to automatically identify and troubleshoot
configuration errors that may occur \cite{xu2015systems}. One such tool,
\cite{keller2008conferr} measures system resilience to configuration
errors by injecting automatically generated configuration errors into the system
and profiling performance.
A test suite can be used to model correct behavior, as in
DirectFix~\cite{mechtaev:directfix}, which formulates program repair as a \maxsmt{} problem
and deems a program fixed when all tests.
In domains that are not amenable to unit-testing user input can be used as a
specification for correction behavior.
For example, Tortoise~\cite{tortoise:weiss} propagates complete system
configuration fixes from a user shell to a system configuration
specification.

\technique{} uses the existing code structure of FSMs to model behaviors without
special procedures for individual behaviors, and thus can repair parameters without
a descriptive model of the robot's behavior. Deterministic test cases cannot
be run on physical robots, so \technique{} leverages user-provided corrections
as a partial specification of how a behavior is incorrect and the desired change.
Since these corrections can be contradictory \maxsmt{} is used to minimize
changes and maximize the number of satisfied corrections, such that resulting
repairs automatically yield the desired behavior while generalizing to other
scenarios.

\subsection{Behavior Synthesis}
In many cases the problem of
configuration repair in robotics can be seen as improving behavior, as
is the focus of \technique{}.
Improving behaviors is a wide
ly studied problem in robotics,
and many modern approaches focus on behavior synthesis, either
by learning complete behaviors, or by synthesizing partial control structure.
Reinforcement learning approaches are a popular method for learning policies
using Markov decision processes.
Examples of reinforcement learning include learning the RoboCup keep-away task
with episodic SMDP Sarsa learning \cite{stone2005keepawayRef}, and
Q-value reuse leveraged for transfer learning of a simpler behavior to
a more complex one using a Nao humanoid robot \cite{stone2010transfer}.
More recently \cite{george2017Param} uses a two-tiered approach which combines
policy search and Q-learning to
learn an action-selection policy with parameterized actions.ce
Hierarchical state machines can be used for reinforcement learning, as in
\cite{bai2017efficient} which uses hierarchies of abstract machines to
short-circuit and speed up Q-learning by identifying internal transitions.


Deep learning techniques have become an extremely popular and promising
approaches for behavior synthesis in recent years \cite{Snderhauf2018TheLA}.
These approaches have been successful in simulated
continuous domains using techniques
such as proximal policy optimization \cite{ppo}, or on real robots for grasping
as in \cite{levine2018handeye} which combines CNNs for success prediction
with a continuous servoing mechanism.
Alternatively, learning action sequences and parameters has been applied in
humanoid robot soccer to teach a full goal scoring policy
using parameterized action spaces \cite{stone2016Rdrein}.
Similar to learning a transition function between tasks
\cite{veloso2018switch} formulates task switching as a Markov decision process
and uses a Dueling Deep Q-Network to find the optimally policy.
This method speeds up execution of a policy by identifying specific
\textit{stimuli} which are significant sensory inputs in the switch decision.

Human input can help overcome the limitations of autonomous
algorithms~\cite{Kamar2016DirectionsIH,nashed2018human}. Learning from
demonstration (LfD)~\cite{argall2009survey} and inverse reinforcement
learning (IRL)~\cite{abbeel2011inverse} allow robots to learn new behaviors
from human demonstrations. LfD can also overcome
model errors by correcting portions of the state
space~\cite{mericli2012multi}. These approaches require demonstrations
in the full high-dimensional state space of the robot,
which cane tedious for users to provide.
When human demonstrations do not specify
\emph{why} an action was applied to a state, it can be hard to generalize
to a new situation. \cite{barto2015grounded} attempts to address
this by using statistical reasoning and control theory to convert
continuous demonstrations to more generalized discrete representations and
modeling multi-step tasks with finite-state representations.
\cite{cakmak2019enduser} uses LfD to approach the problem of goal and action
learning for
new environments using a task specific model for shelf-arrangements,
along with learning strategies that allow varying amounts of human specification for
the task parameters.

Robot behavior can also be repaired by dynamic synthesis of new
control structures and through program synthesis \cite{program-synthesis},
such as automatic synthesis of new
FSMs~\cite{wong2014correct}, synthesis of code from a context-free motion
grammar with parameters derived from human-inspired control~\cite{dantam2013}.
\emph{Programming by example} synthesizes programs from
a small number of examples~\cite{gulwani2011automating} and can also
support noisy data~\cite{devlin:robustfull}.
When automated synthesis is intractable, a user-generated specification
in a domain-specific language can be used to synthesize
a plan~\cite{nedunuri2014}, or to specify high-level behavior using abstractions
such as Instruction Graphs~\cite{mericli2014interactive}.

\technique{} is not intended to synthesize new program structure, but to repair
structurally sound behaviors when their parameters do not reflect the hardware
or environment. The assumption is made that the FSM does not need new structure,
but that failures are due to \emph{incorrect triggering} of the transition function
arising from incorrect environment-dependent parameter values.
The correct values for these parameters are
determined by using corrections as a partial specification of the behavior
in a new environment.
\technique{} works by leveraging the existing structure of the behavior alongside a
small number of samples to provide the minimal parameter adjustments which
satisfy user expectations. This minimal adjustment uses inferred dependencies
from the code to generalize.

\section{Background}
\seclabel{background}

We use a real-world example to motivate \technique: a robot soccer attacker
that
\begin{inparaenum}[1)]
\item goes to the ball if the ball is stopped,
\item intercepts the ball
if it is moving away from the attacker,
\item catches the ball if it is moving toward the attacker, and
\item kicks the ball at the goal once the attacker has control of the ball.
\end{inparaenum}
Each of these sub-behaviors is a distinct, self-contained feedback
controller (\eg{} ball interception~\cite{biswas2014opponent}, two-stage optimal
control \cite{Balaban2018ART}, or
omnidirectional time optimal control~\cite{kalmar2004near}).
At each time-step,
the attacker 1) switches to a new controller if
necessary and 2) invokes the current controller
to produce new outputs. We represent the attacker
as a \emph{robot state machine} (RSM), where each state represents a
controller (\figref{example-model}).

In this paper, we assume that the output of each controller is
nominally correct: there may be minor \emph{performance} degradation
when environmental factors change, but we assume that they are
\emph{convergent}, and will eventually produce the correct result.
Each
controller has preconditions that describe properties of the
robot and world state that should be true when they operate. A
nominally correct controller will eventually produce the desired result if
executed when its preconditions are met. These preconditions are not trivial
to define, and it is the goal of the transition function to approximate them with parameters in order to transfer operation between controllers.
However, environmental factors also affect the transition function.
For example, the friction
coefficient between the ball and the carpet
affects when the attacker transitions from {\itshape Intercept}
to {\itshape Kick}; and the mass of the ball affects when the
attacker transitions from {\itshape Kick} to {\itshape Done}.
These factors vary from one environment to another.
Since transition functions do not have any
self-correcting mechanisms, robots are prone to behaving incorrectly when their
parameters are incorrect for the given environment.

\subsection{Robot State Machines}

A robot state machine (RSM) is a discrete-time Mealy machine that is extended
with continuous inputs, outputs, and program variables. Formally, an RSM is a
9-tuple $\langle S, S_0, S_F, V, V_0, Y, U, T, G\rangle$, where $S$ is the
finite set of states, $S_0 \in S$ is the start state, $S_F \in S$ is the end
state, $V \in \reals{m}$ is the set of program variable values, $V_0 \in
\reals{m}$ are the initial values of the program variables, $Y\in\reals{n}$ are
the continuous inputs, $U\in\reals{l}$ are the continuous actuation outputs, $T
: S \times Y \times V \rightarrow S$ is the transition function, and $G : S
\times Y \times V \rightarrow U \times V$ is the emission function. At each time
step $t$, the RSM first uses the transition function to select a state and
then the emission function to run the controller associated with that state.
The transition function can only update the current state, whereas the emission
function can update program variables and produce outputs.

\subsection{Transition Errors}

In general, transition errors occurs when the values of the parameter
in the transition function do not reflect the environment or properties of the hardware.
A transition error occurs when at a timestep $t$ the output of the transition
function is either: a state representing a controller whose preconditions are not met
and so the controller fails, or a state other than some behavior critical state
with preconditions that are met. Automatically identifying when either type of error
has occurred is a non-trivial problem which does not generalize between RSMs, and
so \technique{} uses human input to identify incorrect triggering of the transition function.

\figref{example-trajectories} shows two traces of the attacker described above. In the
blue trace, the attacker correctly intercepts the moving ball and kicks it at
the
goal. But, in the red trace, the attacker fails to kick: it remains stuck
in the \emph{Intercept} state and never transitions to \emph{Kick}.
Over the course of several trials (\eg{} a robot soccer game), we may find that
the attacker only occasionally fails to kick.
When this occurs, it is usually the case that the high-level structure of
the transition function is correct, but that the values of the
parameters need to be adjusted. Unfortunately, since there are 11 real-valued
parameters in the full attacker RSM, the search space is large.

To efficiently search for new parameter values, we need to reason
about the structure of the transition function. To do so,
the next section describes how we systematically convert
it to a formula in propositional logic, extended
with arithmetic operators. This formula encodes the structure of the
transition function along with constraints from the user corrections, and allows
an SMT solver to efficiently find new parameters to correct the errant
transition(s).

\section{Interactive Robot Transition Repair}
\seclabel{repairable_controllers}

The \technique{} algorithm has four inputs: 1) the transition function,
2) a map from parameters to their values, 3) an
execution trace, and 4) a set of user-provided corrections.
The result of \technique{} is a set of
corrected parameter maps that
each attempt maximize
the number of corrections satisfied and minimizes the changes to the
input parameter map. (The trade-off between these objective is a
hyperparameter.)

\technique{} has three major steps.
\begin{inparaenum}[1)]
  \item A user observes transition errors, identifies their locations in
  the execution trace recorded during operation, and makes a correction
  that specifies the desired alternate transition(s) throughout the trace (\secref{corrections}).
  \item For each user-provided correction,
  \technique{} \emph{partially evaluates}
  the transition function for the inputs and variable
  values at the time of correction, yielding \emph{residual transition functions}
  (\secref{program-analysis}).
  \item Finally, it uses the residual transition functions to formulate an
  optimization problem
  for an off-the-shelf solver that can support first order logic, for the
  majority of this evaluation we use z3, an optimizing \maxsmt{},
  but we also evaluate the performance of dReal, an automated reasoning
  tool for solving first-order logic problems that specializes in nonlinear
  real functions in \secref{smtSolving}. The solution to
  this problem is a set of possible adjustments
  to the parameter values (\secref{max-smt}).
\end{inparaenum}

\begin{figure}
 \centering
 \begin{subfigure}[b]{0.49\linewidth}
    \centering
    \includegraphics[width=0.9\columnwidth]{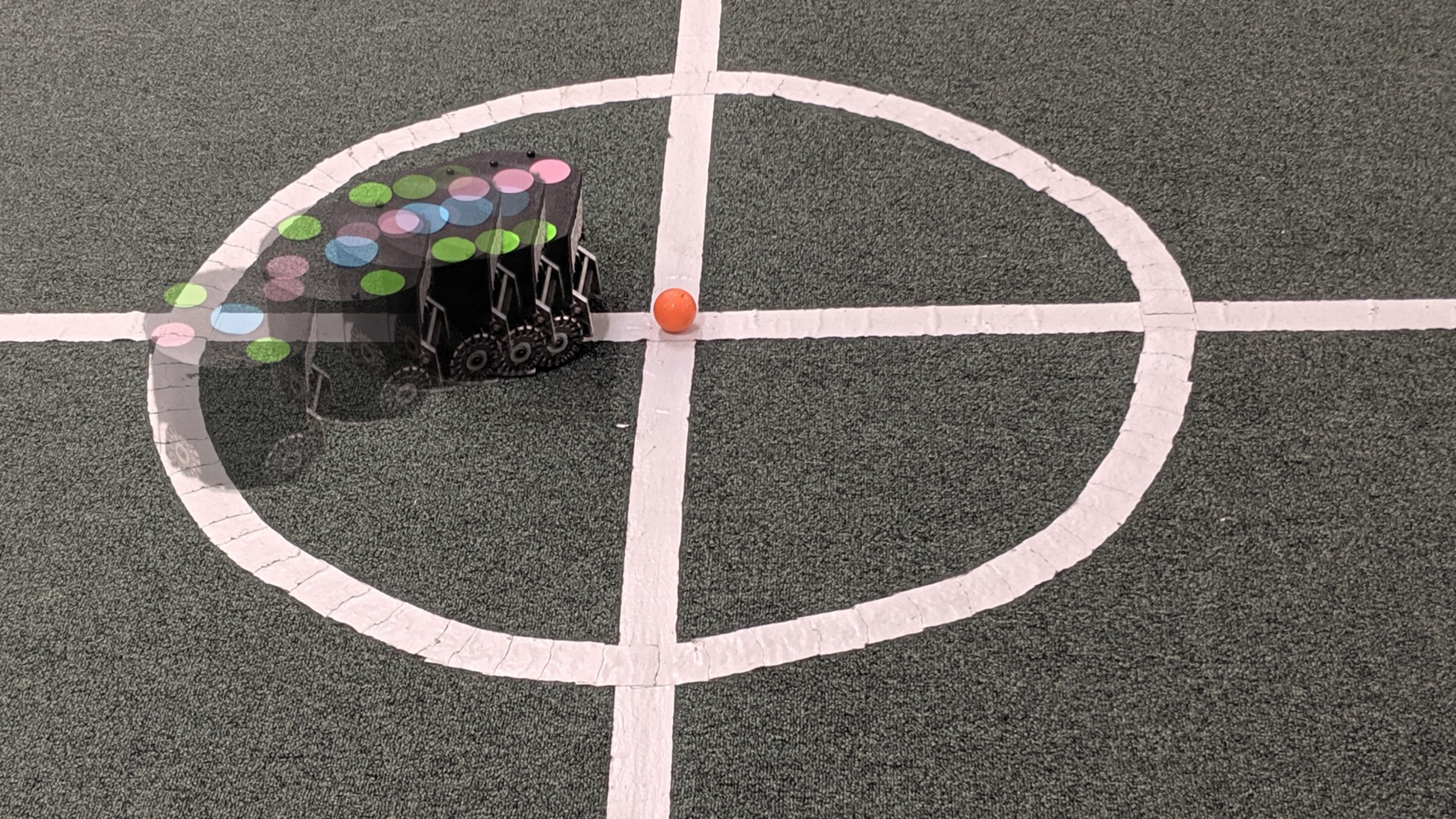}
 \end{subfigure}
 \begin{subfigure}[b]{0.49\linewidth}
    \centering
    \includegraphics[width=0.9\columnwidth]{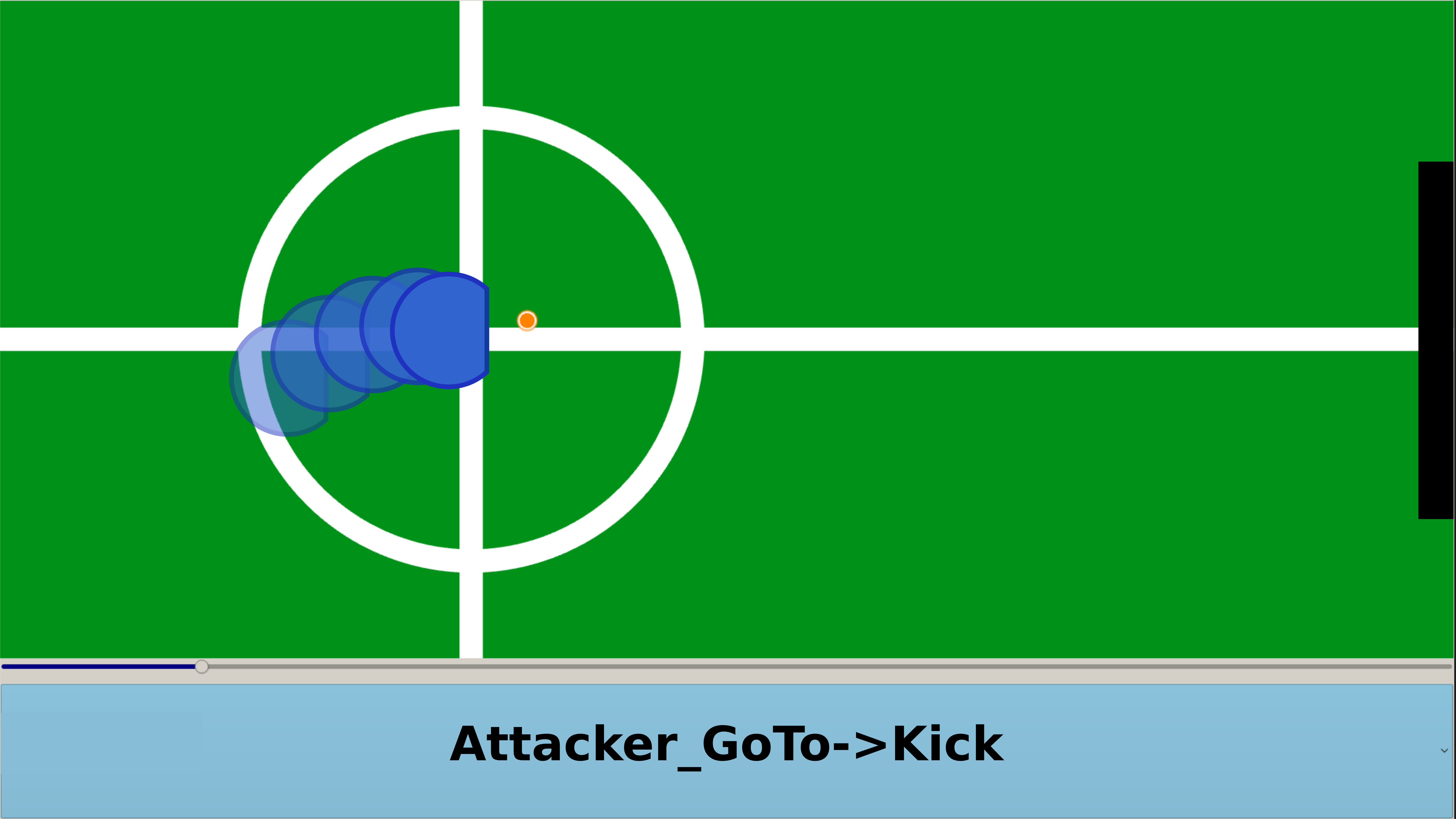}
  \end{subfigure}
  \vspace{-.7em}
  \caption{Example playback of attacker trace where the robot moves into
  a kicking position and fails to transition from GoTo to Kick. The dropdown menu
  shows the user supplied correction. The black bar represents the goal.}
  \figlabel{behavior}
\end{figure}

To illustrate the \technique{} algorithm, we present as a running example a simplified
attacker RSM that is only capable of handling
a stationary ball on the field (\figref{simple-rsm}).
Therefore, the RSM has four
states (Start, GoTo, Kick, and End) and its transition function
(\figref{example-function}) has
four parameters (\cpparam{aimMargin}, \cpparam{maxDist}, \cpparam{viewAng}, and
\cpparam{kickTimeout}). We run the RSM with initial parameter values and observe
the behavior shown in \figref{behavior} which shows an example execution and playback
trace where the attacker fails to transition.
In the execution trace from the RSM
(\figref{example-inputs}) a user identifies that at time-step $t=5$ the transition
function produces an incorrect result (GoTo), and provides the desired output (Kick)
as a correction in \figref{behavior}.

The goal of \technique{} is to find an
adjustment to the parameters such that the transition function produces the
corrected next state instead of the recorded state at time $t+1$.
With this example in mind, we present how \technique{} uses the transition
function code, an execution trace, and a correction to
identify that an adjustment to just one of the parameters, \cpparam{maxDist}, is sufficient
to satisfy this correction (\figref{example-outputs}).

\begin{figure}
  \centering
  \begin{tikzpicture}[scale=0.05]
  \tikzstyle{every node}+=[rounded corners,minimum width=0.9cm,minimum
  height=0.35cm,node distance=1.5cm and 1.7cm,draw=black,inner
  sep=0pt,font=\tiny\itshape]
  \tikzstyle{every path}+=[>=latex]
  \node[fill=blue!25] (start) {Start};
  \node (goto) [right of=start] {Go To};
  \node (kick) [right of=goto] {Kick};
  \node[fill=green!25] (end) [right of=kick] {End};
  \path[every loop/.style={looseness=15}]
  (start) [->] edge (goto)
  (goto) [->] edge [loop below] (goto)
  (goto) [->] edge (kick)
  (kick) [->] edge [loop below] (kick)
  (kick) [->] edge (end);
  \end{tikzpicture}
  \caption{Simplified attacker RSM.}
  \figlabel{simple-rsm}
\end{figure}
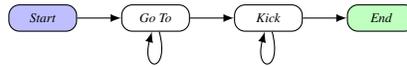

\subsection{Transition Functions and \technique{} Inputs}
\seclabel{syntax}

To abstract away language-specific details of our repair procedure, we present
\technique{} for an idealized imperative language that only has features
essential for writing transition functions.  \figref{syntax} lists the
syntax for a  transition function written in \technique{}-repairable form using
a notation that is close to standard BNF.  In
general, the transition function consists of sequences of statements comprised
of expressions over 1) the current state \pstate{}, 2) program inputs \pin{x},
3) parameters \pparam{x}, and 4) program variables \pvar{x}. Based on
computations over these identifiers, the transition function returns the next
state.  \figref{syntax} has a list of operators that often appear in transition
functions, such as arithmetic and trigonometric functions, but the list is
\emph{not} exhaustive. As a concrete example of a transition function written in
repairable form, \figref{example-function} shows the transition function
for the running example: it branches on the current state (\pstate) and returns
the next state. The crux of the transition function are the conditions that
determine when the transition from \state{Go To} to \state{Kick} occur.

A parameter map ($P$) specifies the parameters of a transition function.
\figref{example-inputs} shows the parameter map of our example before running
\technique{}. The output of \technique{} will be a set of adjustments to this
parameter map.

An execution trace is a sequence of \emph{trace elements} $\trelt_t$. A
trace element records the values of sensor inputs, program variables, and the
state at the start of time-step $t$. Finally, a user-provided \emph{correction}
($\correlt{}$) specifies the expected state at the end of time-step $t$, and
optionally a set of parameters $U$ to consider for repair, if no $U$ is
provided, all parameters will be considered for repair.
In our example, the attacker should have transitioned to the \state{Kick}
state at $t=5$.
\figref{example-inputs} shows the trace element and a user correction for the
running example: since the correction $c_5$ refers only to the time-step $t=5$,
only the relevant trace element $\trelt_5$ is shown.

\subsection{Providing Corrections}
\seclabel{corrections}
Repairs with \technique{} require human input for identifying what went wrong
during RSM execution. In order to facilitate the best repairs, \technique{}
supports two interaction models for identifying corrections. Each
correction technique is useful for identifying corrections with
respect to different error types. We call these correction techniques
\textit{\imm{s}} and \textit{\continue{s}}.

Immediate corrections are made using the exact states in the execution trace, the user identifies
a timestep $t$ in the trace when an error occurred, the desired output
state $s_d$, and optionally a set of parameters to repair $U$. These
corrections allow the user to consider all of the world states that are explored
in an execution trace, and to specify behavior modification
based on exact world and program states that resulted in errors.
Allowing the user to specify $U$ gives some flexibility to the specificity of a repair.
When specific parameters are known to be wrong identifying them apriori
can simplify the
repair process and yield more controlled results, when the incorrect parameters
are unknown \technique{} will identify the appropriate parameters to repair
based on the desired transitions. Immediate
corrections are effective in general when the user can isolate states where
a desired transition \textit{should} occur, and many \imm{s}
can be combined for more complex repairs.

However, it is sometimes the case that a user identifies a state with
a premature transition into a state $s_e$,
and the desired behavior is to remain
in the prior state $s_p$. In many such cases a \textit{should} transition
correction with respect to state $s^p$ is equivalent to a \textit{should not}
transition correction with respect to $s^e$. A single correction of this
form results in repairs that either do not fully specify the transition
function output, or that yield minimal changes. Since individual
controllers converge on a desired world state, the most common result of
using \imm{s} in these scenarios is to modify
the behavior for a single timestep before allowing the undesirable transition
again. This repair output
is desirable if the erroneous transition was only a single
timestep early, but in many cases the desired behavior is actually
for the RSM to \textit{continue} without transitioning to
$s^e$ for a number of timesteps.

To illustrate this, consider a case of the attacker
transitioning to kick prematurely as in \figref{early_kick}.
In this example the attacker transitions well before the
desired preconditions for kick are met, there is no point in the trace
where the correct world state for the transition is observed. However, giving
a correction of the form \textit{do not transition to kick} at the timestep
where this transition occurs will yield the behavior shown in \figref{still-early}.
The difference in the two world states can be hard to see, because the end
result is that the attacker \textit{continues} the GoTo state for one
timestep longer than it did prior to the repair. Using \imm{s}
 we would need to repair and test continuously in order to
reach the desired preconditions for kick.

\begin{figure}
  \begin{adjustwidth}{-3.0cm}{-3.0cm}
 \centering
 \begin{subfigure}[b]{0.3\linewidth}
  \centering
  \includegraphics[width=0.5\linewidth]{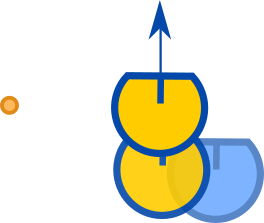}
  \caption{Premature kick.}
  \figlabel{early_kick}
 \end{subfigure} \hspace{-3.em}
 \begin{subfigure}[b]{0.3\linewidth}
 \centering
  \includegraphics[width=0.45\linewidth]{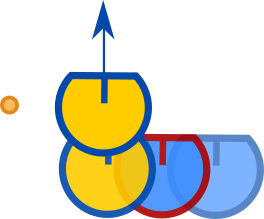}
  \caption{One negative correction.}
  \figlabel{still-early}
 \end{subfigure}
 \begin{subfigure}[b]{0.3\linewidth}
 \centering
  \includegraphics[width=1.0\linewidth]{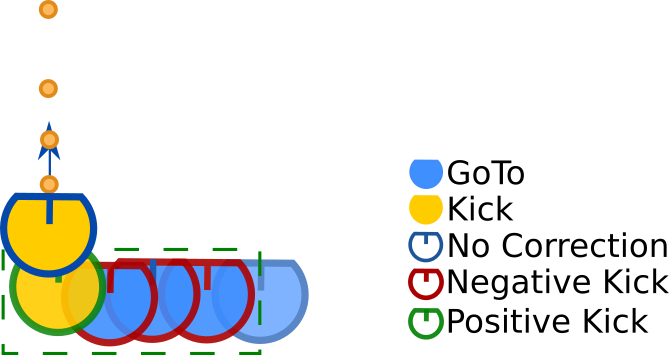}
  \caption{Continue correction}
  \figlabel{continue-trace}
 \end{subfigure}
 \caption{Example traces of attacker execution illustrating continue corrections. Ball is in
 orange, robot
 states are represented by the fill, and the outline labels when a correction was applied.
 The green box shows the set of corrections that result from a \continue{}.}
  \figlabel{both_too_early}
  \end{adjustwidth}
\end{figure}

A user signals a \continue{} by identifying a trace element $\tau_t$,
the state \texttt{s}$^c$ that was incorrectly output at $\tau_t$,
and optionally a set of parameters to repair $U$. This
signals the users desire to continue execution from the state described
in $\tau_t$ while \textit{not} transitioning to \texttt{s}$^c$.
However, the trace will not contain future timesteps where
the observed transition did not occur, and so new trace elements
need to be generated. To do this \technique{} forks the world state
by using $\tau_t$ to calculate an initial world state and
RSM configuration for forward simulation. The RSM configuration
prevents the
transition function from transitioning to \texttt{s}$^c$ and during simulation
a negative correction is generated for each timestep until the
user identifies a timestep $n$ where the behavior should transition to \texttt{s}$^c$.
The end result
is a set of negative constraints $D = \{c_t,...,c_{t+n}\}$, a single
positive constraint $c_{t+n}::= $\texttt{s}$_{t+n+1}^c$, and corresponding $U$
for all constraints. Specifying $U$ can be particularly useful for \continue{s}
because the negative constraints only need to invalidate one subclause
of a transition clause to prevent a transition, as opposed to positive
constraints that must satisfy the entire clause. In this case the specification
of $U$ can guide the solver towards specific subclauses, which may improve
generalization in some cases.

\figref{continue-trace} visualizes an example execution trace after forking
the world state in
a \continue{} for the attacker RSM. Robot outlines in red represent locations
where a negative constraint was created, while the green outline shows the
single positive kick constraint. The green box shows all of the corrections
generated in the forked world state of the \continue{}. In this
example four corrections
were generated with a \continue{}
that otherwise would have required iterative repair after
each correction to yield the same result with \imm{s}.

Using immediate and continue corrections together allows a user to
specify immediate transition points present in an execution trace, as well
as to create corrections that delay premature transitions until a desired state
is reached. A user may supply corrections using either or both techniques,
and then repair with \technique{}.
Additionally, if the behavior is functioning correctly in some scenarios and
not in others, then the user can use execution traces of desirable behavior to
provide a set of $n$ corrections $N = \{c_i,...,c_{n}\}$ that are representative
of nominal transition behavior. In this case the supplied $c_i$ do not specify
alternative transitions, but instead reinforce existing transition behavior.
Providing $N$ alongside a number of corrections to erroneous transitions
adds constraints that
directs the solver to find repairs that fix the error cases if possible, but that
equally attempts to maintain the nominal transitions behavior
described by $N$.

The end result of either correction technique
is a set of corrections $C = \{c_t,...,c_{t+n}\}$ that specifies the desired
alternative behavior that should result from the final \technique{} repair.
In the case of our simple running example we have a single \imm{}
$c_5$, as shown in \figref{example-inputs}.

\begin{figure}
  \begin{adjustwidth}{-5cm}{-5cm}
  \centering
  \begin{subfigure}[b]{0.32\paperwidth}
  \lstset{language=srtr-example}
  \scriptsize
  \scriptsize
  \textbf{Transition function} ($T$)
  \begin{lstlisting}
  if ($\pstate$ == "START") {
    return "GOTO";
  } else if ($\pstate$ == "GOTO") { #\label{line:goto-start}#
    $\cpid{relLoc}$ := $\cpin{ballLoc} - \cpin{robotLoc}$;  #\label{line:relLoc}#
    $\cpid{aimErr}$ := $\textrm{AngleMod}(\cpin{targetAng} - \cpin{robotAng})$;
    $\cpid{robotYAxis}$ := $\langle \sin(\cpin{robotAng}), \psub\!\cos(\cpin{robotAng})\rangle$; #\label{line:trig-input}#
    $\cpid{relLocY}$ := $\cpid{robotYAxis} \cdot \cpid{relLoc}$; #\label{line:yLoc}#
    $\cpid{maxYLoc}$ := $\cpparam{maxDist} \cdot \sin(\cpparam{viewAng})$; #\label{line:maxYLoc}#  #\label{line:trig-param}#
    if ($\cpid{aimErr} < \cpid{aimMargin} \wedge \left\lVert\cpid{relLoc}\right\rVert < \cpparam{maxDist} \wedge$  #\label{line:kick-guard-start}#
        $\left\lVert\cpid{relLocY}\right\rVert < \cpid{maxYLoc} \wedge$
        $\cpin{time} > \cpvar{lastKick} + \cpparam{kickTimeout}$) {  #\label{line:kick-guard-end}#
      return "KICK";
    } else return "GOTO"; #\label{line:goto-end}#
  } else if ($\pstate == "KICK" \wedge $
             $\cpvar{timeInKick} > \cpparam{kickTimeout}$) {
    return "END";
  } else return "KICK";
  \end{lstlisting}
  \caption{A simple RSM and its transition function.}
  \figlabel{example-function}
  \end{subfigure}
  \vrule\;
  \begin{subfigure}[b]{0.2\paperwidth}
  \lstset{language=srtr-example}
  \scriptsize
  \(
  \begin{array}{@{}r@{\;}c@{\;}l}
  \multicolumn{3}{@{}l}{\textbf{Parameter map ($P$)}} \\
  P & = & \langle \cpparam{aimMargin} \pmapsto \pi/50, \\
  & & \phantom{\langle} \cpparam{maxDist} \pmapsto 80, \\
  & & \phantom{\langle} \cpparam{viewAng} \pmapsto \pi/6, \\
  & & \phantom{\langle} \cpparam{kickTimeout} \pmapsto 2\rangle\\[0.7em]
  \hline
  \multicolumn{3}{@{}l}{\textbf{Trace element ($\trelt_5$)}} \\
  \treltins{\trelt_5} & = &
    \langle \cpin{ballLoc} \pmapsto \langle 30,40 \rangle, \\
    & & \phantom{\langle} \cpin{robotLoc} \pmapsto \langle 0,0 \rangle, \\
    & & \phantom{\langle} \cpin{robotAng} \pmapsto 0, \\
    & & \phantom{\langle} \cpin{targetAng} \pmapsto \pi/60, \\
    & & \phantom{\langle} \cpin{time}\pmapsto 5\rangle\\
  \treltvars{\trelt_5} & = & \langle \cpvar{lastKick} \pmapsto 2 , \\
    & & \phantom{\langle} \cpvar{timeInKick}\pmapsto 0\rangle \\
  \treltstate{\trelt_5} & = & \texttt{"GOTO"} \\[0.7em]
  \hline
  \multicolumn{3}{@{}l}{\textbf{Trace ($\mathcal{R}$)}} \\
  \mathcal{R} & = & \langle \cdots \tau_5 \cdots \rangle \\[0.7em]
  \hline
  \multicolumn{3}{@{}l}{\textbf{Correction ($c_5$)}} \\
  c_5 & ::= &  s_6 \pmapsto \texttt{"KICK"}
  \end{array}
  \)
  \caption{Inputs to \technique{}.}
  \figlabel{example-inputs}
  \end{subfigure}
  \;
  \vrule
  \;
  \begin{subfigure}[b]{0.3\paperwidth}
  \lstset{language=srtr-example}
  \scriptsize

  \textbf{Repairable and unrepairable parameters}

  \(
  \begin{array}{@{}r@{\,}c@{\,}l}
  \textsf{Rep}(T) & = & \{ \cpparam{aimMargin}, \cpparam{maxDist},
    \cpparam{kickTimeout} \} \\
  \textsf{Unrep}(T) & = & \{ \cpparam{viewAng} \}
  \end{array}
  \)

  \vskip 0.5em
  \hrule
  \vskip 0.2em
  \textbf{Result of } $\textsf{MakeResidual}(T,\tau_5,P)$
  \vskip -0.7em
  \begin{lstlisting}
  if ($\pi/60 < \cpparam{aimMargin} \wedge  50 < \cpparam{maxDist} \wedge$
      $40 < \cpparam{maxDist} \cdot 0.5 \wedge 5 > 2 + \cpparam{kickTimeout}$) {
    return "KICK";
  } else return "GOTO";
  \end{lstlisting}

  \vskip -0.5em
  \hrule
  \vskip 0.2em
  \textbf{Result of} $\textrm{CorrectOne}(T,\tau_5,P,c_5)$:

  \(
  \begin{array}{@{}r@{\,}c@{\,}l}
  \phi  & = & \exists \delta^1,\delta^2,\delta^3 : \pi/60 < \pi/50+\delta^1 \wedge 50 < 80+\delta^2 \wedge \\
  & & \phantom{\exists\delta^1,\delta^2,\delta^3 : } 40 < (80 + \delta^2) \cdot 0.5 \wedge 5 > 2 + (2 + \delta^3)\\
  \end{array}
  \)

  \vskip 0.5em
  \hrule
  \vskip 0.2em
  \textbf{Result of} $\textrm{CorrectAll}(T,P,\mathcal{R},\{c_5\})$:

  \(
  \begin{array}{@{}r@{\,}c@{\,}l}
  \Phi &  = & \exists \delta^1,\delta^2,\delta^3,w^1 : w^1 = H \veebar (w^1 = 0 \wedge \phi)
  \end{array}
  \)

  \vskip 0.5em
  \hrule
  \vskip 0.2em
  \textbf{Result of} $\textrm{SRTR}(T,P, 1, \mathcal{R},\{c_5\})$ for $H = 1$:

  \(
  \begin{array}{@{}r@{\,}c@{\,}l}
  & & \underset{w^1,\delta^{1\cdots 3}}{\mathrm{arg\,min}} w^1 + \| \delta^1\| + \| \delta^2\| + \| \delta^3\| ~\textrm{constrained by}~\Phi \\
  & = & \langle w^1 = 0, \delta^1 \mapsto 0, \delta^2 \mapsto 0.5, \delta^3 \mapsto 0 \rangle \\
  \end{array}
  \)
  \caption{Each step of the \technique{} algorithm.}
  \figlabel{example-outputs}
  \end{subfigure}
  \caption{\technique{} applied to a simplified robot soccer attacker with a single correction.}
  \end{adjustwidth}
  \end{figure}

  \begin{figure}
  \centering
  \scriptsize
  \lstset{language=model}
  \begin{minipage}[t]{0.3\columnwidth}
  \(
  \begin{array}{@{}r@{\;}c@{\;}ll}
  \multicolumn{4}{@{}l}{\textbf{Unary Operators}} \\
  \mathit{op}_1 & ::= & \multicolumn{2}{l}{\texttt{-}
    \mid \texttt{sin}
    \mid \texttt{cos}
    \mid \cdots} \\[.4em]
  \multicolumn{4}{@{}l}{\textbf{Expressions}} \\
  \expr & ::=  & k & \hspace{20pt} \textrm{Constants} \\
    & \mid & \pstate  & \hspace{20pt} \textrm{State} \\
    & \mid & \pvar{x} & \hspace{20pt} \textrm{Variables} \\
    & \mid & \pin{x} & \hspace{20pt} \textrm{Inputs} \\
    & \mid & \pparam{x} & \hspace{20pt} \textrm{Parameters} \\
    & \mid & \multicolumn{2}{l}{\hspace{-6pt} \mathit{op}_1(\expr)} \\
    & \mid & \multicolumn{2}{l}{\hspace{-6pt} \expr_1~\mathit{op}_2~\expr_2}
  \\[.4em]
  \multicolumn{4}{@{}l}{\textbf{Transition Functions}} \\
  T & ::=  & \multicolumn{2}{l}{\texttt{\{ $\stmt_1$; $\cdots$; $\stmt_n$ \}}}
  \end{array}
  \)
  \end{minipage}
  \vrule
  \hspace{1pt}
  \begin{minipage}[t]{0.3\columnwidth}
  \(
  \begin{array}{@{}r@{\;}c@{\;}ll}
  \multicolumn{4}{@{}l}{\textbf{Binary Operators}} \\
  \mathit{op}_2 & ::= & \texttt{+}
    \mid \texttt{-}
    \mid \texttt{*}
    \mid \texttt{>}
    \mid \cdots \\[.4em]
  \multicolumn{4}{@{}l}{\textbf{Statements}} \\
  \stmt & ::=  & \textbf{return}~\pstate\texttt{;} \hspace{42pt} \textrm{Return the next state s} \\
    & \mid & \pvar{x}~\texttt{\passign}~\expr\texttt{;} \hspace{45pt} \textrm{Update \pvar{x} to e} \\
    & \mid & \textbf{\texttt{if}}~(\expr)~\stmt_1~\textbf{\texttt{else}}~\stmt_2
    \hspace{10pt} \textrm{Conditional} \\
    & \mid & \texttt{\{ $m_1\cdots m_n$ \}} \hspace{20pt} \textrm{Statement block} \\[.4em]
  \multicolumn{4}{@{}l}{\textbf{Parameter Maps}} \\
  P & ::= & \multicolumn{2}{l}{\langle \pparam{x}^1\pmapsto k^1\cdots
  \pparam{x}^n\pmapsto k^n\rangle} \\[.4em]
  \multicolumn{4}{@{}l}{\textbf{Designated Parameters}} \\
  U & ::= & \multicolumn{2}{l}{\{ \pparam{x}^i \in P \}} \textrm{Subset of P to repair.}\\[.4em]
  \multicolumn{4}{@{}l}{\textbf{Traces}} \\
  \ptrace & ::= & \lbrack \trelt_1 \cdots \trelt_n \rbrack \\[.4em]
  \multicolumn{4}{@{}l}{\textbf{Corrections}} \\
  \correlt & ::= &  \pstate \in S, U ::= \{x_p^1\cdots x_p^k\}
  \end{array}
  \)
  \end{minipage}
  \vspace{0.4em}
  \hrule
  \vspace{0.4em}
  \(
  \begin{array}{@{}r@{\;}c@{\;}ll}
  \multicolumn{4}{@{}l}{\textbf{Trace Elements}} \\
  \trelt_t & ::= & \multicolumn{2}{@{}l}{\mktrelt{
    \langle \limforall{i=1}{m} \pin{x^i} \pmapsto k^i \rangle}
    {\langle \limforall{j=1}{n} \pvar{x^j}\pmapsto k'^j \rangle}
    {\pstate_t \pmapsto k''_s}} \\[.4em]
  \end{array}
  \)
  \caption{Syntax of transition functions, traces, and corrections.}
  \figlabel{syntax}
  \end{figure}

  \begin{figure}[t]
  \begin{center}
  \begin{adjustwidth}{-3.0cm}{-3.0cm}
  \lstset{language=algo}
  \begin{lstlisting}[multicols=2]
    // Takes a transition function $T$, and returns a partially evaluated residual transition
    // function $T'$ by eliminating identifiers $x^i$ using their values $k^i$.
    def Peval($T$,$x^1\pmapsto k^1\cdots x^n\pmapsto k^n$); #\label{line:peval}#

    // Returns the list of repairable parameters of the transition function $T$
    def Rep($T$);

    // Returns the list of unrepairable parameters of the transition function $T$
    def Unrep($T$);

    def Residual($T$,$\trelt_t$,$P$,$U$):#\label{line:residual}#
      $\mktrelt{
      \langle \limforall{i=1}{l} \pin{x^i} \pmapsto k^i \rangle}{
      \langle \limforall{j=1}{m} \pvar{x^j} \pmapsto k'^j \rangle}{
      \pstate_t \pmapsto k''_s}$ := $\trelt_t$
      $\{ \pparam{x^1},\cdots,\pparam{x^n}\}$ = Unrep($T$, $U$)#\label{line:residual-unrep}#
      $T'$ := Peval($T$,$\limforall{i=1}{l} \pin{x^i} \pmapsto k^i,
       \limforall{j=1}{m} \pvar{x^j}\pmapsto k'^j,
       \limforall{k=1}{n} \pparam{x^k}\pmapsto P(\pparam{x^k}),
       \pstate_t \pmapsto k''_s$)#\label{line:residual-input-spec}#
      return $T'$#\label{line:residual-end}#

    def CorrectOne($T$,$\trelt_t$,$P$,$U$, $\correlt$):#\label{line:correctone-start}#
      $T'$ := Residual($T$,$\trelt_t$, $P$, $U$)#\label{line:correctone:residual}#
      $\{ \pparam{x^1},\cdots,\pparam{x^m}\}$ := Rep($T$)#\label{line:correctone:rep}#
      return $\exists \delta^1,\cdots,\delta^m:\correlt = T'(
    \pstate_1,\pparam{x'^1} +
    \delta^1,\cdots,\pparam{x'^m}+\delta^m)$#\label{line:correctone-end}#

    def CorrectAll($T$, $P$, $U$, $\ptrace$, $\{\correlt^1, \cdots, \correlt^n\}$):
      $\{ \pparam{x^1},\cdots,\pparam{x^m}\}$ = Rep($T$)
      $\Phi$ = $\mathrm{true}$
      for $i \in [1\cdots n]$:
        $\exists \delta^1, \cdots, \delta^m : \phi_i$ = CorrectOne($T$,$\ptrace[t]$,$P$,$U$$\correlt^i$)
        $\Phi$ = $\Phi \wedge (w^i = \phyper \veebar (w^i = 0 \wedge \phi_i))$ #\label{line:correctall:xor}#
      return $\exists \delta^1,\cdots,\delta^m,w^1,\cdots,w^n : \Phi$

    def SRTR($T$, $P$, $U$, $k$, $\ptrace$,$\{\correlt^1, \cdots, \correlt^n\}$): #\label{line:srtr-start}#
      $\Phi$ = CorrectAll($T$,$P$,$U$,$\ptrace$,$\{\correlt^1, \cdots, \correlt^n\}$))
      $S$ = {}
      for $i \in [1\cdots k]$:
        assert($\Phi$)
        minimize($\Sigma_{i=1} w^i + \Sigma^m_{j=1} \|\delta^j\| : \Phi$) #\label{line:srtr:minimize}#
        $r = \langle w_i \in \{0,H\}, \delta^j \mapsto a^j \rangle$ #\label{line:srtr:solution}#
        $S = S \cup r$ #\label{line:srtr:union}#
        $\Phi = \Phi \land (\exists w^i : w^i = 0 \land w^i \in r \neq 0)$ #\label{line:srtr:notsame0}#
        $\Phi = \Phi \land (\exists w^i : w^i = H \land w^i \in r \neq H)$ #\label{line:srtr:notsameH}#
      return $S$ #\label{line:srtr-end}#
  \end{lstlisting}
  \caption{The core \technique{} algorithm.}
  \figlabel{srtr-algo}
\end{adjustwidth}
\end{center}
\end{figure}

\subsection{Residual Transition Functions}
\seclabel{program-analysis}
\lstset{language=algo}

For each correction ($c_t$)
\technique{} needs to translate the transition function and $c_t$ into a formula
for an SMT solver. To do this \technique{} first
simplifies the problem by specializing the transition function using
the state, variables, and inputs recorded at time $t$.
We
call this simplified transition function the \emph{residual transition function}. A
residual transition function consists only of real numbers and expressions
containing parameters, or only those containing the parameters in $U$
if one $U$ is specified, and
considers only the branch of the transition function corresponding
to $c_t$. This is achieved by substituting the input and variable
identifiers with concrete values from the trace element and simplifying expressions
as much as possible using an approach known as \emph{partial evaluation}~\cite{jones:pe}.
\figref{example-outputs} shows the output of \lstinline|MakeResidual| for
our example correction at $t=5$.
Since the
state at this time-step (\treltstate{\tau_5}) is \state{Go To},
the residual transition function only has the code from the
branch that handles this case (\ie{} the code from lines
\ref{line:goto-start}--\ref{line:goto-end}) and contains only the information
\technique{} will need to translate the residual transition function into
a formula for an SMT solver.

A potential problem with this approach is that many solvers do not have
decision procedures that support nonlinear arithmetic such as trigonometric functions,
which occur frequently in RSMs. Our example also
uses trigonometric functions in several expressions. Fortunately, most of these
trigonometric functions are applied to inputs and variables, thus they
are substituted with concrete values in the residual. For example,
line~\ref{line:trig-input} calculates
\lstinline|$\sin(\cpin{robotAng})$| and \lstinline|$\cos(\cpin{robotAng})$|,
but \lstinline|$\cpin{robotAng}$| is an input. Thus, the residual substitutes
the identifier with its value from the trace element
($\trelt_5.\cpin{robotAng}= 0$) and simplifies the trigonometric
expressions. In contrast, line~\ref{line:trig-param} applies a trigonometric
function to a parameter (\lstinline|$\sin(\cpparam{viewAng})$|).
This makes \lstinline|$\cpparam{viewAng}$| an
\emph{unrepairable parameter} for many solvers that cannot appear in the
residual transition function. \technique{} substitutes unrepairable parameters
with their concrete values from the parameter map when the backend solver
does not support them.

In general, the \lstinline|MakeResidual| function of \technique{}
(lines~\ref{line:residual}--\ref{line:residual-end} in \figref{srtr-algo})
takes a transition function ($T$), a trace element ($\tau_t$), a parameter
map ($P$), and a set of possible parameters to repair ($U$) and produces a residual transition function by partially
evaluating the transition function with respect to the trace element and
the unrepairable parameters. We use a simple dataflow analysis to
calculate the unrepairable parameters (\lstinline|Unrep($T$)|)
and a canonical partial evaluator (\lstinline|Peval|)~\cite{jones:pe}.

\subsection{Transition Repair as a \maxsmt{} Problem}
\seclabel{max-smt}

Given the procedure for calculating residual transition functions, \technique{}
proceeds in three steps. 1) It translates each correction $c_i$
into an independent formula $\phi_i$. A solution to $\phi_i$ corresponds
to parameter adjustments that satisfy the correction $c_i$.
Note however that no solution exists if $c_i$ cannot be satisfied. 2) It
combines the formulas $\phi_{1\cdots n}$ for all corrections $c_{1\cdots n}$
from the previous step into a single
formula $\Phi$ with independent penalties $w^i$ for each sub-formula $\phi_i$.
A solution to $\Phi$ corresponds to parameter adjustments that satisfy a subset of the
corrections. Any unsatisfiable corrections incur a penalty. 3)
Finally, it formulates a \maxsmt{} problem that minimizes the magnitude of
adjustments ($\|\delta^j\|$) to the parameters, and the penalty ($w^i$) of
violated sub-formulas.

The \lstinline|CorrectOne| function transforms a single correction into a
formula.
This function
1) calculates the residual transition
function (\figref{srtr-algo}, line~\ref{line:correctone:residual}), 2) gets the
repairable
parameters (line~\ref{line:correctone:rep}) and,
and 3) produces
a formula (line~\ref{line:correctone-end}) with a variable for each
repairable parameter in $U$ for the current correction.
In our running example, the transition function has four parameters and no $U$
is specified,
but, as explained in the previous section, the residual has only
three parameters since \cpparam{viewAng} is unrepairable.
Therefore, the formula that corresponds to this residual (\figref{example-outputs})
has three variables ($\delta^1$, $\delta^2$, and $\delta^3$).
Moreover, since the correction ($c_5$) requires the next-state to be
\state{Kick}, which only occurs when the residual takes the true-branch
(line 3 of the residual), the body of the formula is equivalent to
the conditional expression (lines 1--2), but with each parameter replaced
by the sum of its concrete value (from $P$) and its adjustment (a $\delta$).
For example, the formula replaces $\cpparam{aimMargin}$
by $\pi/50 + \delta^1$. Therefore, when $\delta^1 = 0$, the parameter
is unchanged.

The \lstinline|CorrectAll| function supports multiple corrections
and uses \lstinline|CorrectOne| as a subroutine.
 The function iteratively builds
a  conjunctive formula $\Phi$, where each clause has a distinct
penalty $w^i$ and two mutually exclusive cases: either
$w^i = 0$ thus the clause has no penalty and the adjustments to the parameters
satisfy the $i$th correction (line~\ref{line:correctall:xor}); or
a penalty is incurred ($w^i = \phyper$) and the $i$th correction is violated.
Thus $\phyper \in \reals{+}$ is a hyperparameter that induces a trade-off between
satisfying
more corrections \vs{} minimizing the magnitude of the adjustments:
large values of $\phyper$ satisfy more corrections with larger adjustments,
whereas small values of $\phyper$ satisfy fewer corrections with smaller adjustments.
The final formula has $m$ real-valued variables $\delta^i$ for adjustments to
the corresponding $m$ repairable parameters $\pparam{x}^i$, and
$n$ discrete variables $w^j$ that represent the penalty of violating formula
$\phi_j$ corresponding to correction $\correlt^j$.

Our example (\figref{example-outputs}) has one correction and three repairable
parameters. Therefore,
\lstinline|CorrectAll|
produces a formula with four variables: a single penalty ($w^1$)
and the three adjustments discussed above ($\delta^{1}$, $\delta^{2}$,
and $\delta^{3}$). The formula is a single exclusive-or: either
the penalty is zero and formula is equivalent to the result of
\lstinline|CorrectOne| or the penalty is one and the result of
\lstinline|CorrectOne| is ignored.

\subsection{\maxsmt{} Solutions}
\seclabel{solving}
In order to solve for adjustments to parameters
the  function \lstinline|SRTR| uses \lstinline|CorrectAll| as a
subroutine and invokes the \maxsmt{} solver. This function
1) stores the formula returned by \lstinline|CorrectAll| as $\Phi$,
2) initializes an empty solution set $S$,
3) and iteratively solves and updates a version of the \maxsmt{} problem
$k$ times as specified by the user.
To solve the \maxsmt{} problem at each iteration the current problem
$\Phi$ is asserted, and then \lstinline|SRTR| directs the solver to minimize
the sum of the penalties and the sum of the magnitude of parameter changes
(line~\ref{line:srtr:minimize} in \figref{srtr-algo}). The resulting
weights $w^i$ and real-valued adjustments $a^j$ to the $\delta^j$ are stored
as $r$, and added to the solution set $S$.

At this point in \lstinline|SRTR| we have a single solution $r \in S$ that
represents a possible adjustment to the parameters that satisfies
some subset of the user corrections.
Applying this to our running example we can see that
the minimum-cost solution in the first iteration has
$\delta^2 = 0.5$ with other variables set to zero.
\ie{} we can satisfy the correction by adjusting \cpparam{maxDist}
from $80$ to $80.5$. In many cases this initial solution will be the best solution
to the problem, and in cases such as our example, the only solution.

However, when there are multiple corrections and it is not possible or desirable
to satisfy all user corrections there may be many partially satisfying solutions.
The goal of \technique{} is to capture user intent accurately by identifying the best repair
out of the possible solutions, and this is done in part by iteratively solving
updated version of the \maxsmt{} problem in \lstinline|SRTR|.
The single iteration \maxsmt{} approach, also the first solution in $S$,
assumes that the best solution is the one which satisfies the largest
number of corrections while minimizing the parameter adjustments. Adjusting
the value of the hyperparameter $H$ adjusts the weighting between minimizing
adjustments and maximizing satisfied corrections, and uses the same weight
for all corrections. This method is sensible when there is not a priority
ranking between corrections, or when all corrections aim to achieve similar
repairs to the behavior.

Alternatively, in the case where there is clear user
preference between
supplied corrections, a separate weighting system can be used that gives
corrections individual weights according to their priority,
such that more important corrections are more likely to be satisfied.
However, it is infrequently known apriori what the relative weights between
all constraints should be, and so it is difficult for a user to accurately
weight corrections as they are made.

An alternative to these apriori weights is to instead utilize the solver to explore
other possible models iteratively, through a process called \textit{model enumeration.}
Model enumeration for SMT works by adding a constraint to the problem after
each iteration of solutions that excludes the previous model from the possible
valid solutions. For our problem we consider a constraint system with $n$
corrections as discussed in \secref{max-smt}:
\begin{align}
  \eqlabel{smtForm}
  \Phi =\exists \{ \delta^1 \cdots \delta^m \},
  \{ w^1 \cdots w^n \} : w^i = H \veebar (w^i = 0 \wedge \phi^i),\
  \min{} \sum_{i=0}^n w^i + \sum_{j=0}^m ||\delta^j||
\end{align}
After one iteration the solver returns a solution of the form:
\begin{align}
  \eqlabel{smtSolution}
  r  = \langle w^i= 0 \vee H, \delta^j \mapsto a^j \rangle
\end{align}
for some real number adjustments to parameters $a^j$.
In order to enumerate a new model constraint we would search for a new
solution $r'$ under the new constraint $r' \neq r$.
This type of model enumeration works well for SAT problems, and SMT problems
that are not optimizing an objective function. However, for our formulation this
type of model enumeration is not sufficient to generate meaningfully different
models. The fact that our objective functions
seeks to minimize the real valued $\delta^i$s gives the solver room to make
minimal real valued changes that do not yield meaningfully different final models.

To force the system to find novel solutions with conflicting parameter
sets \technique{} implements a type of
solution exploration in lines $36-39$ in \figref{srtr-algo}.
At each iteration a new solution $r$ is calculated by the solver and
$\Phi$ is updated with a two-part constraint
that prohibits
the previous solution and forces solutions with novel changes at the correction level.
Given $\Phi$ and $r$ the first constraint requires that some correction
that was previously satisfied
should not be satisfied in the new solution, (line~\ref{line:srtr:notsame0} in \figref{srtr-algo}).
This alone can allow an optimization that only satisfies fewer constraints,
so we also add a constraint that requires some previously unsatisfied correction
to be satisfied, (line~\ref{line:srtr:notsameH} in \figref{srtr-algo}).
Using these additional constraints we can search for new models iteratively by
adding further restrictions after each new solution, up to some iteration limit,
or until all possible models have been explored.
The end result is a set of solutions $S$ and the satisfied corrections
for each repair which can be evaluated as needed by the user.

In summary, SRTR adjusts parameters to satisfy user provided corrections.
It is not always possible to find an
adjustment that satisfies all corrections. Moreover, there is
a tradeoff between making larger adjustments and satisfying
more corrections. Therefore \technique{} uses a \maxsmt{} model to formulate
parameter adjustment. This model captures the tradeoffs between satisfying
correction and minimizing adjustments, and allows for the exploration of different
solutions when not all corrections are satisfiable.

\subsection{Alternate Solvers}
\seclabel{smtSolving}
We formulate the optimization problem used by \technique{} as a \maxsmt{} problem
that attempts to minimize the adjustments to parameters while maximizing the number
of corrections satisfied.
This formulation can then be fed to an off-the-shelf solver, and different
solvers offer different potential advantages.
In this article we evaluate the performance of two solvers with
\technique{} by comparing Z3 \cite{de2008z3}, an SMT theorem prover,
and dReal \cite{dreal} an automated reasoning tool
for solving problems encoded as first-order logic formulas over the real numbers.
We have chosen these two solvers because Z3 is a popular SMT solver,
and because dReal
has potential advantages for our applications in robotics.

In particular, dReal specializes
in handling problems that involve nonlinear real functions, such as
trigonometric functions, which could reduce the amount of partial evaluation
needed, and allow \technique{} to repair parameters even when they are used
in nonlinear functions. To evaluate dReal with nonlinear functions
we considered the single \technique{} correction shown here:
\begin{align*}
\phi  =  \exists \delta^1: 10.0 < 20 * sin(\pi+\delta^1)
\end{align*}
In this simple example we have a case of a tuneable parameters ($\delta^1$) inside
a trigonometric function, where a solution is simple,
adjust $\delta^1$ by roughly $-.5$.
A \maxsmt{} solver such as Z3 cannot find a solution to this problem, but
dReal quickly finds a minimal change of $\delta^1=-0.524$.
However, since dReal is not an SMT solver, it is not designed
for \maxsmt{}, and so does not support soft-constraints.
The \maxsmt{} formulation is a major component of
\technique{} as it enables \continue{s}, solution exploration, and
conflicting correction sets in a straightforward manner. In order to test
dReals capability to handle these properties we modeled the \maxsmt{} optimization
problem described in \secref{max-smt}
using first-order logic and the single correction $\phi$ to generate the following
problem:
\begin{align*}
\Phi = \exists \delta^1,w^1 : w^1 = H \veebar (w^1 = 0 \wedge \phi)
\end{align*}
Given this new formulation the solver now has the option to optimize between
satisfying the correction, and minimizing the cost. For the same $\phi$
we would expect the correction to only be unsatisfied if the value of $H$ is
greater than the magnitude of the adjustment needed to repair.
Since here we repair with a $H > 1.0$ the expected outcome is the same
as before. Despite this,
the result from dReal is neither optimal, nor satisfies
the constraint, as it suggests a repair of
$\delta^1=2886.58$.

This result suggests that even
formulated in first order logic, the full \maxsmt{} formulation of \technique{}
cannot be used with dReal. However, the core of \technique{} can be modeled
in dReal, and so dReal is an alternative solution when nonlinear arithmetic
is more prevalent than the need for solution exploration and \maxsmt{}. We further
evaluate the performance of dReal, alongside z3 in \secref{performance}.

\section{Evaluation}

In this section we evaluate \technique{} using four RSMs and their respective success
rates as follows.
The \emph{attacker} (\figref{example-model}) fills the main offensive role in
robot soccer. Its success
rate is the fraction of the test scenarios where it successfully
kicks the ball into the goal. The \emph{deflector} (\figref{deflection-fsm})
plays a supporting role in robot soccer, performing one-touch
passing~\cite{bruce2008cmdragons}. Its success rate is
the fraction of the test scenarios where it successfully deflects
the ball. The \emph{docker} (\figref{docking_fsm}) is a non-soccer behavior which drives a
differential drive robot to line up and dock with a charging station. Its success rate is the fraction of the test scenarios where
it successfully docks with the charging station. Finally, the \emph{passing} behavior
(\figref{passing_fsm}) is a simple autonomous car behavior which attempts to move
through slower traffic in a safe manner. The success rate for passing is the fraction
of the test scenarios where it successfully passes all slower cars without coming
too close to any other vehicle in the process.
We use these RSMs in a number of experiments to evaluate:
\begin{enumerate}
  \item How \technique{} compares to exhaustive search;
  \item The solution time and success rate of repairs with two different
  solver as the number of correction used varies;
  \item The affect of \imm{s} on RSM success rate as
  compared to the effects of expert tuning;
  \item Continue corrections, techniques for improving generalization, and their performance
  with respect to premature \imm{s};
  \item How solution exploration quantitatively and qualitatively affects the
  performance of the attacker RSM; and finally
  \item If \technique{} can be used to improve the performance of a real-world
  competitive soccer robot.
\end{enumerate}

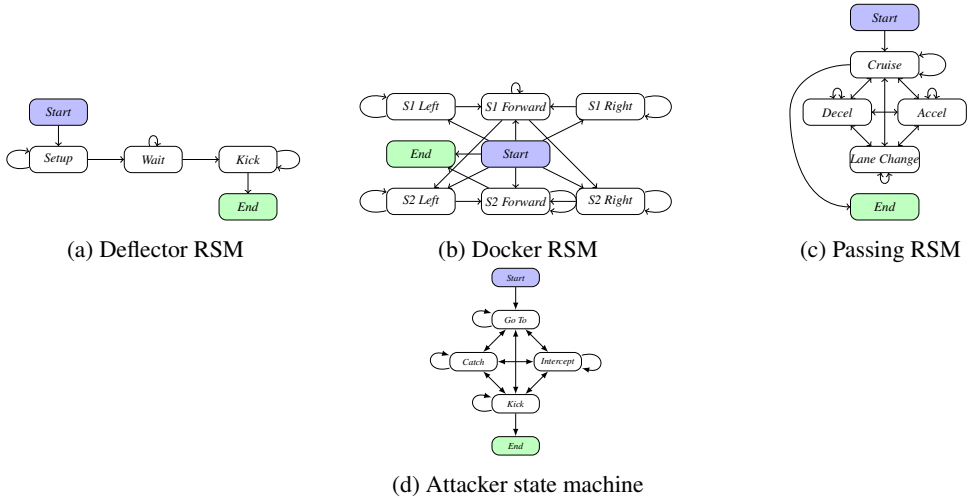
\begin{figure}
  \begin{adjustwidth}{-10.0cm}{-10.0cm}
  \centering
  \begin{subfigure}[b]{0.15\linewidth}
   \centering
   \scalebox{.7}{%
    \begin{tikzpicture}[scale=0.151]
      \tikzstyle{every node}+=[rounded corners,minimum width=1.1cm,minimum height=0.5cm,node distance=0.9cm and 2cm,draw=black,inner sep=0pt,font=\scriptsize\itshape]
      \node[fill=blue!25] (start) {Start};
      \coordinate [right of=start] (rightstart);
      \node (setup) [below of=start] {Setup};
      \coordinate [right of=setup] (rightsetup);
      \node (wait) [right of=rightsetup] {Wait};
      \coordinate [right of=wait] (rightwait);
      \node (kick) [right of=rightwait] {Kick};
      \coordinate [right of=kick] (rightkick);
      \coordinate [left of=kick] (leftkick);
      \node[fill=green!25] (end) [below of=kick] {End};
      \path[every loop/.style={looseness=5}]
        (start) [->] edge (setup)
        (setup) [->] edge (wait)
        (wait) [->] edge (kick)
        (kick) [->] edge (end)
        (setup) [->] edge [loop left] (setup)
        (wait) [->] edge [loop above] (wait)
        (kick) [->] edge [loop right] (kick);
     \end{tikzpicture}
   }
   \caption{Deflector RSM}
  \figlabel{deflection-fsm}
  \end{subfigure}
  \begin{subfigure}[b]{0.15\linewidth}
  \centering
    \scalebox{.7}{%
    \begin{tikzpicture}[scale=0.151]
      \tikzstyle{every node}+=[rounded corners, minimum width=1.3cm,minimum height=0.5cm,node distance=0.9cm and 2cm,draw=black, inner sep=0pt,font=\scriptsize\itshape]
      \node[fill=blue!25] (start) {Start};
      \coordinate [left of=start] (leftstart);
      \node (s1forward) [above of=start] {S1 Forward};
      \coordinate [left of=s1forward] (left1forward);
      \node (s1left) [left of=left1forward] {S1 Left};
      \coordinate [right of=s1forward] (right1forward);
      \node (s1right) [right of=right1forward] {S1 Right};
      \node (s2forward) [below of=start] {S2 Forward};
      \coordinate [right of=s2forward] (right2forward);
      \coordinate [left of=s2forward] (left2forward);
      \node (s2left) [left of=left2forward] {S2 Left};
      \node (s2right) [right of=right2forward] {S2 Right};
      \node[fill=green!25] (end) [left of=leftstart] {End};
      \path[every loop/.style={looseness=5}]
        (start) [->] edge (s1forward)
        (start) [->] edge (s1left)
        (start) [->] edge (s1right)
        (start) [->] edge (s2forward)
        (start) [->] edge (s2left)
        (start) [->] edge (s2right)
        (start) [->] edge (end)
        (s1left) [->] edge (s1forward)
        (s1right) [->] edge (s1forward)
        (s1forward) [->] edge (s2left)
        (s1forward) [->] edge (s2right)
        (s2left) [->] edge (s2forward)
        (s2right) [->] edge (s2forward)
        (s2forward) [->] edge (end)
        (s1left) [->] edge [loop left] (s1left)
        (s2left) [->] edge [loop left] (s2left)
        (s1forward) [->] edge [loop above] (s2forward)
        (s2forward) [->] edge [loop right] (s2forward)
        (s1right) [->] edge [loop right] (s1right)
        (s2right) [->] edge [loop right] (s2right);
    \end{tikzpicture}
    }
  \caption{Docker RSM}
  \figlabel{docking_fsm}
  \end{subfigure}%
  \begin{subfigure}[b]{0.15\linewidth}
  \centering
     \scalebox{.7}{%
      \begin{tikzpicture}[scale=0.151]
        \tikzstyle{every node}+=[rounded corners, minimum width=1.3cm,minimum height=0.5cm,node distance=0.9cm and 2cm,draw=black, inner sep=0pt,font=\scriptsize\itshape]
        \node[fill=blue!25] (start) {Start};
        \node (cruise) [below of=start] {Cruise};
        \coordinate [below of=cruise] (belowcruise);
        \node (decel) [left of=belowcruise] {Decel};
        \node (accel) [right of=belowcruise] {Accel};
        \coordinate [right of=accel] (rightaccel);
        \coordinate [left of=decel] (leftdecel);
        \node (switch) [below of=belowcruise] {Lane Change};
        \node[fill=green!25] (end) [below of=switch] {End};

        \path[every loop/.style={looseness=5}]
          (start) [->] edge (cruise)
          (decel) [<->] edge (accel)
          (decel) [<->] edge (switch)
          (accel) [<->] edge (switch)
          (cruise) [<->] edge (decel)
          (cruise) [<->] edge (accel)
          (cruise) [<->] edge (switch)
          (cruise) edge [loop right] (cruise)
          (decel) edge [loop above] (decel)
          (accel) edge [loop above] (accel)
          (switch) edge [loop below] (switch);
        \draw [<-] (end) to [out=180,in=-90] (leftdecel) to [out=90, in=180](cruise);
      \end{tikzpicture}
      }
  \caption{Passing RSM}
  \figlabel{passing_fsm}
  \end{subfigure}%

  \begin{subfigure}[b]{0.15\linewidth}
  \centering
    \scalebox{.7}{%
    \begin{tikzpicture}[scale=0.151]
      \tikzstyle{every node}+=[rounded corners,minimum width=0.9cm,minimum
      height=0.35cm,node distance=0.8cm and 1.7cm,draw=black,inner
      sep=0pt,font=\tiny\itshape]
      \tikzstyle{every path}+=[>=latex]
      \node[fill=blue!25] (start) {Start};
      \node (goto) [below of=start] {Go To};
      \coordinate[below of=goto] (belowgoto);
      \node (intercept) [right of=belowgoto] {Intercept};
      \node (catch) [left of=belowgoto] {Catch};
      \node (kick) [below of=belowgoto] {Kick};
      \node[fill=green!25] (end) [below of=kick] {End};
      \path[every loop/.style={looseness=5}]
        (start) [->] edge (goto)
        (goto) [<->] edge (kick)
        (goto) [<->] edge (intercept)
        (goto) [<->] edge (catch)
        (intercept) [<->] edge (kick)
        (catch) [<->] edge (kick)
        (catch) [<->] edge (intercept)
        (kick) [->] edge (end)
        (goto) [->] edge [loop left] (goto)
        (catch) [->] edge [loop left] (catch)
        (intercept) [->] edge [loop right] (intercept)
        (kick) [->] edge [loop left] (kick);
    \end{tikzpicture}
    }
  \caption{Attacker state machine}
  \end{subfigure}%

  \caption{RSMs used for experiments.}
  \figlabel{example_RSMs}
  \end{adjustwidth}
\end{figure}

\subsection{Comparison To Exhaustive Search}
\seclabel{parameter_search}

Using the attacker, we compare \technique{} to an exhaustive search to show
that 1) \technique{} is dramatically faster and 2) the adjustments found by
\technique{} are as good as those found by exhaustive search.
To limit the cost of exhaustive search, the experiment only repairs the
six parameters that affect transitions into the \state{Kick} state;
we bound the search space by the physical
limits of the parameters; and we discretize the resulting hypercube in
parameter space. We evaluate each parameter set using 13
simulated positions and manually specify if the position should
transition to the \state{Kick} state.

We evaluate the initial parameter values, the \technique{}-adjusted
parameters, and the parameters found by exhaustive search on
20,000 randomly generated scenarios. \tabref{comparison-results} reports
the success rate and running time of each approach. \technique{}
and exhaustive search achieve a comparable success rate. However,
\technique{} completes in 10 ms whereas exhaustive search takes 1,300 CPU
hours (using 100 cores).

\begin{table}
  \centering
  \setlength\tabcolsep{1.5pt} 
  \begin{tabular}{|l| r| c |}
  \hline
  \textbf{Method} &\textbf{Success Rate ($\%$)}  & \textbf{CPU Time} \\
  \hline
  Initial Parameters & 44 & ---  \\
  \hline
  Exhaustive Search & 89 & \hfill 1,300 hr \\
  \hline
  \technique{} & 89 & \hfill 10 ms \\
  \hline
  \end{tabular}
  \vspace{-.5em}
  \caption{Success rate and CPU time compared to exhaustive search.}
  \tablabel{comparison-results}
  \vspace{-.8em}
\end{table}

\subsection{Performance}
\seclabel{performance}

Using the attacker, we evaluate how the number of corrections affects \technique{}
performance in terms of time to solution and success rate while comparing
two different
backend solvers, dReal, and Z3. We evaluate the performance of the attacker in
simulation by discretizing a simulated soccer field into discrete x,y positions,
and starting the attacker at the center of the field.
For each x,y position we set the initial velocity of the ball to a fixed speed,
and vary the angle between $10$ uniformly distributed angles. For evaluating performance we start with an initial parameter
configuration that yields a $\sim 30$\% success rate on these trials, and
generate a set of $150$ correction using failures taken randomly from this dataset. To generate these corrections we employ a nominal parameter
configuration that is successful on a large set of test cases. For each
test case we generate an execution trace using both the nominal parameters,
and a failing parameter set. We identify a correction at the first point
where the two traces diverge, such that the corrected state
is the state in the nominal trace element. This trace element represents the first case
where the poorly performing transition function output a different state
than the nominal configuration.
Since Z3 cannot repair parameters used within nonlinear functions, and dReal
does not support the \maxsmt{} formulation used for handling conflicting corrections,
only corrections with no conflict and no unsatisfiable parameters are used, such
that all combinations are solvable with either solver.

We then sample a test dataset of $1000$ trials from $100$ positions across the field
not used for corrections. Each trial applies \technique{} to a subset of the
corrections, solves the resulting formulation with both Z3 and dReal, and
evaluates the performance on the test dataset. We repeat this procedure 30
times for each number of corrections $N \in [1,30]$, selecting $N$ random corrections
at each iteration, for a total of $900,000$ total trials.

We show the success rate for each solver in \figref{comparison_success}.
It is possible for a single informative correction to dramatically increase the
success rate, or for a particularly under-informative correction to have little
effect. Therefore we also report the mean success rate and show the $99\%$
confidence interval in gray. Both graphs show that even a small number of
corrections can be sufficient for repair, but they also show a general trend
of performance improvement as more corrections are used up to $\sim25$ corrections.
Up to this point both solvers yield comparable final repair performance, with
a peak success rate of $87\%$, a substantial improvement over the initial $30\%$,
however with $> 20$ corrections dReal repair
success rate starts to degrade slightly, to $84\%$ at the lowest. In terms
of success rate both solvers yield significant performance improvement when
applicable, but when large numbers of corrections are used, Z3 is preferable.

\begin{figure}
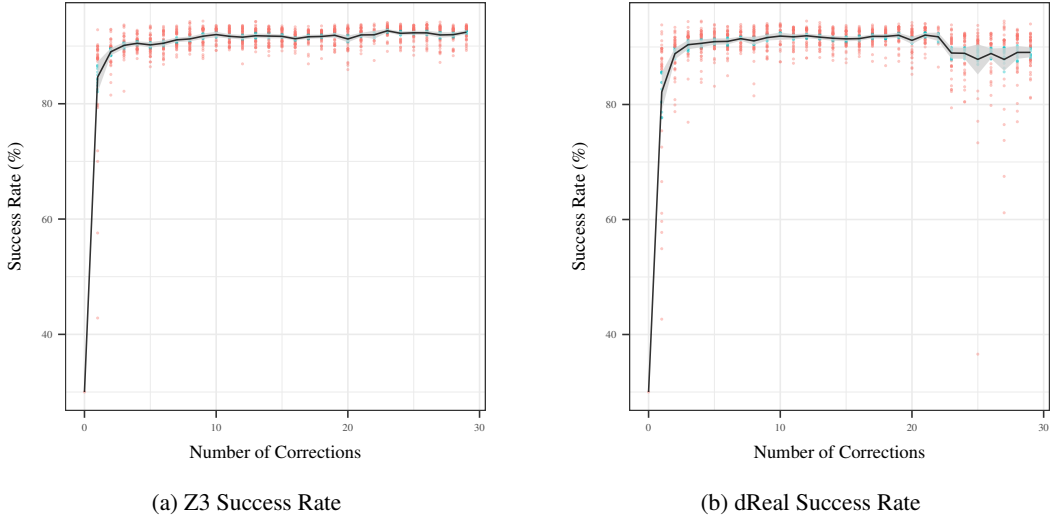

\begin{adjustwidth}{-1.5cm}{-1.5cm}
 \captionsetup[subfigure]{aboveskip=-1pt,belowskip=-4pt}
 \centering
 \begin{subfigure}[b]{0.49\linewidth}
    \centering
    \include{images/z3_corrections_vs_success}
    \caption{Z3 Success Rate}
 \end{subfigure}
 \begin{subfigure}[b]{0.49\linewidth}
    \centering
    \include{images/dreal_corrections_vs_success}
    \caption{dReal Success Rate}
 \end{subfigure}
 \vspace{-.2em}
 \caption{Success rate with different solvers and numbers of corrections. We report the
 mean as a line, the $99\%$ confidence interval in grey, inliers in blue,
 and outliers in red. Darker points
 represent more occurrences.}
 \vspace{-.3em}
\figlabel{comparison_success}
\end{adjustwidth}
 \end{figure}

For evaluating solver time we use the same correction and test datasets as for
success rate, and evaluate using values of $N \in [1,40]$.
The solution times for each
solver are show in \figref{comparison_time}.
Using Z3 solution time increases linearly with the number
of corrections, and the variance in the time taken is relatively small
as the corrections used vary, with $N = 40$ z3 solution time is less than
$.03$ seconds. In contrast, dReal solution time is highly dependent on the
correction set used, and for $N = 40$ varies between $.03$ seconds and $ > 750$
seconds, with an average performance around $250$ seconds. In terms of time
to solution Z3 is preferable to dReal for our use case on average,
with less execution time variance.

\begin{figure}
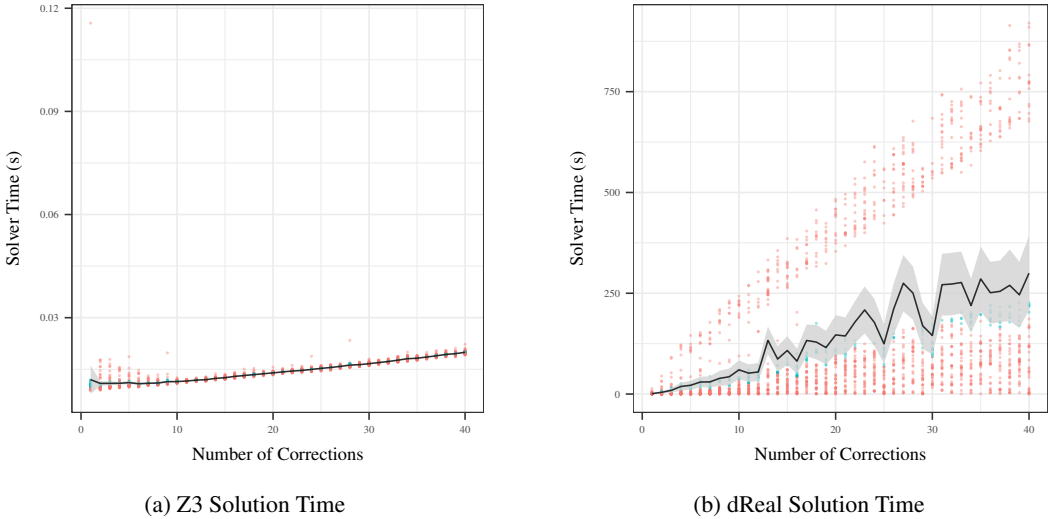

\begin{adjustwidth}{-1.5cm}{-1.5cm}
 \captionsetup[subfigure]{aboveskip=-1pt,belowskip=-4pt}
 \centering
 \begin{subfigure}[b]{0.49\linewidth}
    \centering
    \include{images/z3_vs_time}
    \caption{Z3 Solution Time}
 \end{subfigure}
 \begin{subfigure}[b]{0.49\linewidth}
    \centering
    \include{images/dreal_vs_time}
    \caption{dReal Solution Time}
 \end{subfigure}
 \caption{Solver time with different solvers and numbers of corrections. We report the
 mean as a line, the $99\%$ confidence interval in grey, inliers in blue,
 and outliers in red. Darker points
 represent more occurrences.}
 \vspace{-.3em}
 \figlabel{comparison_time}
 \end{adjustwidth}
\end{figure}

In general this experiment shows that the core of
\technique{} can be used with different backend solvers,
each with different advantages. For the majority of this paper we evaluate
with Z3 because it supports the \maxsmt{} formulation that \technique{} makes
use of, but \technique{} can be translated to other solvers. As such,
in the future \technique{} can expect to leverage emerging technologies and
advances in first-order logic solvers.

\subsection{Objective Tradeoff}
\seclabel{objective_tradeoff}

We evaluate the effect of the objective tradeoff parameter $H$ using Z3. The value
of this parameter determines the weight given to satisfying user corrections
with respect to the magnitude of the adjustment that must be made to satisfy them.
The higher the value of $H$ the greater an adjustment to parameters can
be and still be acceptable for satisfying a correction.
Lower values will favor smaller adjustments to the parameter over satisfying the corrections.

In testing this tradeoff we also test two optimization models, pareto
and lexicographic optimization. Pareto optimization seeks to optimize all of
the objectives simultaneously, while lexicographic optimizes each objective
in sequence, treating the earlier objectives as constraints on the later ones.
We use pareto optimization for all cases where
we desire a weighting between the two parameters, and lexicographic optimization
for when we want to satisfy as many corrections as possible regardless of parameter
adjustments. We do not test the case where we minimize adjustments lexicographically
above maximizing corrections, as this would always result in no repair regardless
of corrections used.

For this test we vary the weighting between lexicographically optimizing in
favor of satisfied corrections, and pareto optimization with a value range
for $H$ of $0.1-10.0$ increasing by $0.1$.
We test by using $10$ sets of $30$ corrections for the attacker RSM, and
we show the number of satisfied corrections and sum of the
parameter percent changes
in  \figref{objective_tradeoff}.
This graph demonstrates that the parameter adjustment magnitudes, and the
number of satisfied correction both increase as $H$ increases.

We show in \figref{comparison_success} that success rate generally increases as the number of user corrections increases. This is by design, as in \technique{} the
user acts as an oracle describing the proper behavior for the RSM. In that
case lexicographic optimization, or values of $H$ favoring satisfying corrections
are generally preferable for success rate. In terms of performance we
graph solver time against the number of corrections
using pareto optimization in \figref{objective_time},
which show that in general lexicographic optimization is preferable to pareto
optimization in terms of solution speed.

\begin{figure}
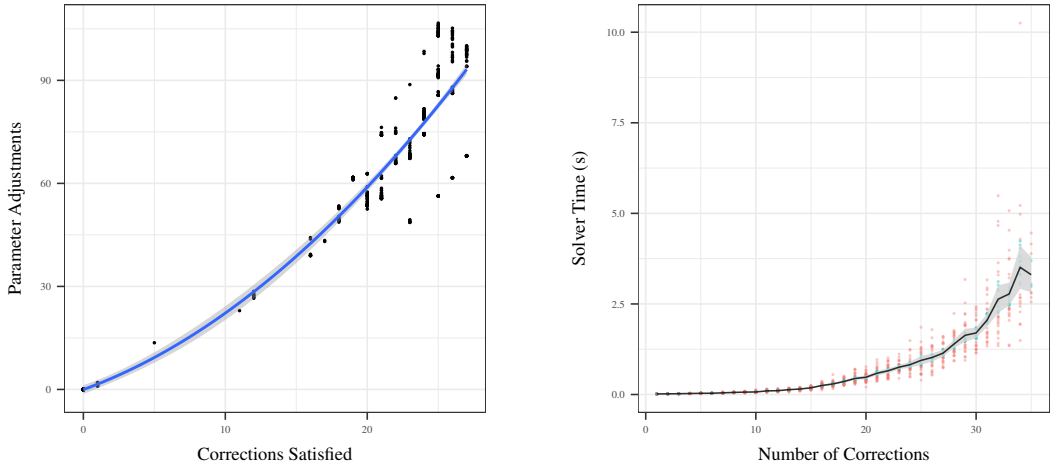

\begin{adjustwidth}{-1.5cm}{-1.5cm}
  \centering
  \begin{subfigure}[b]{0.49\linewidth}
    \captionsetup{justification=centering}
      \centering
    \include{images/objective_tuning}
    \caption{Correlation between parameter percent change, and
    corrections satisfied.}
    \figlabel{objective_tradeoff}
  \end{subfigure}
  \begin{subfigure}[b]{0.49\linewidth}
    \captionsetup{justification=centering}
    \centering
    \include{images/objective_time}
    \caption{Solver time with pareto optimization and varying numbers of
    corrections.}
    \figlabel{objective_time}
  \end{subfigure}

 \caption{\technique{} performance with pareto optimization and varying
 values of $H$.}
\end{adjustwidth}
\end{figure}

\subsection{Immediate Corrections}
\seclabel{controller_repair}

Immediate corrections are one of two correction methods used by \technique{}.
These corrections are used when the desired transition preconditions are
recorded in an execution trace such that an exact world state for a correction
can be identified.
We use four RSMs to show
that when \imm{s} are used
1) \technique{}-adjusted parameters generalize to new scenarios and that 2)
\technique{} outperforms a domain-expert who has 30 minutes to manually
adjust parameters.

\tabref{RSM_stats} summarizes the results of this experiment. We evaluate the
success rate of the Attacker, Deflector, Docker, and Passing RSMs on
test datasets with
several thousand test scenarios each. The baseline parameters that we use for
these RSMs have a low success rate. We give a domain expert complete access to
the RSM code (\ie{} the transition and emission functions), and subsequently
our simulator for 30 mins. In that time, the expert is able to dramatically
increase the success rate of the Deflector, moderately increase the passing
performance, but has minimal impact on the success
rate of the Attacker and the Docker. Finally, we apply \technique{} using a
handful of corrections and the baseline parameters. The
\technique{}-adjusted parameters perform significantly better than the baseline
and domain-expert parameters.

The heat maps in \figref{heatmap} illustrates how parameters found by the domain-expert,
and by \technique{} generalize to novel scenarios with the Attacker. In
both heat maps, the goal is the green bar and the initial position of the
Attacker is at the origin. Each coordinate corresponds to an initial position
of the ball and for each position we set the ball's initial velocity to 12
uniformly distributed
angles. With the expert-adjusted parameters, the Attacker performs well when
the ball starts in its immediate vicinity, but performs poorly otherwise.
However, with \technique{}-adjusted parameters, the Attacker is able to catch
or intercept the ball from most positions on the field.
For this result, we required only two corrections and the cross-marks in the
figure show the initial position of the ball for both corrections.
Therefore, although \technique{} only adjusted parameters to account for these two
corrections, the result generalized to many other positions on the field.

\begin{figure}
 \centering
 \begin{subfigure}[b]{0.49\linewidth}
    \centering
    \includegraphics[width=0.9\columnwidth]{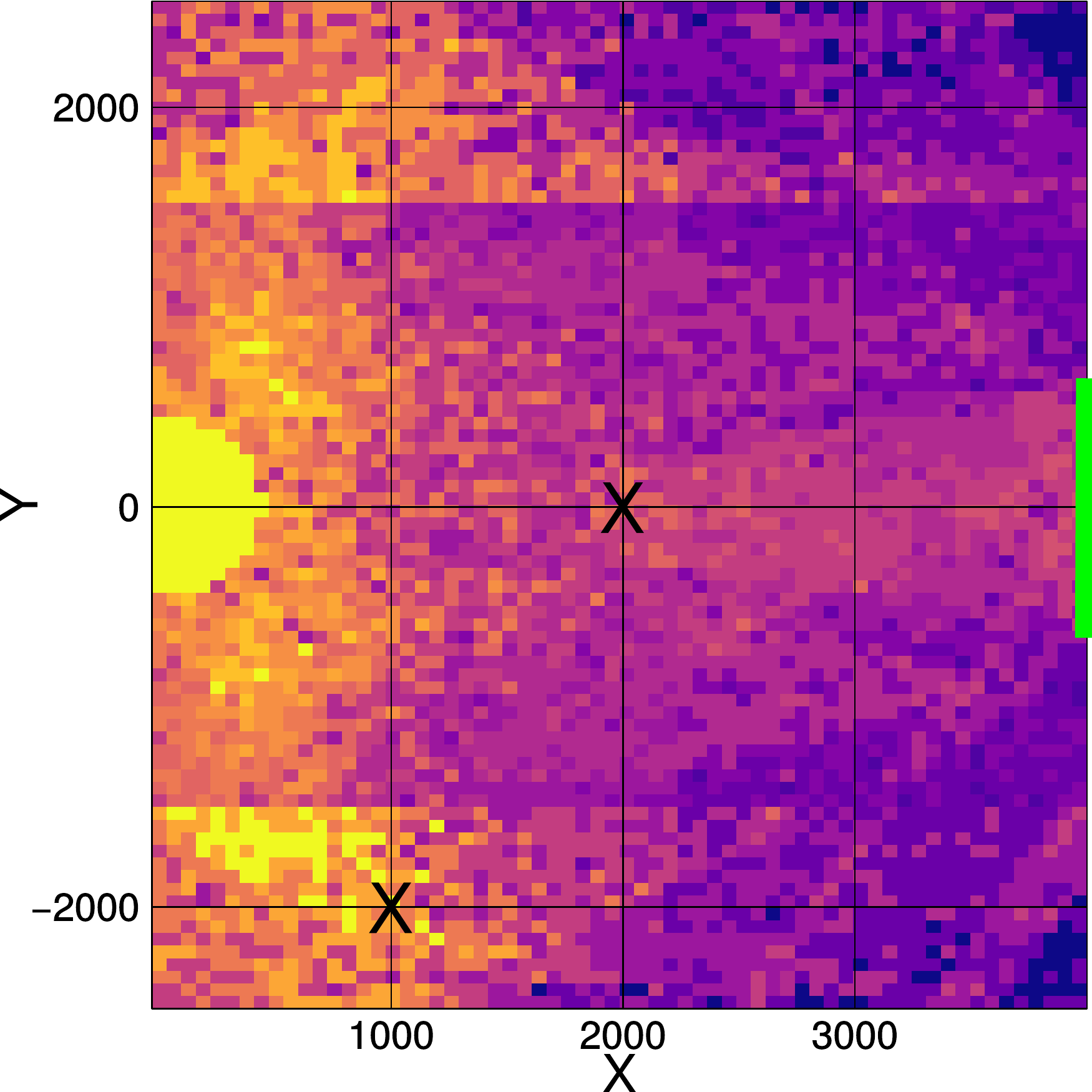}
    \caption{Expert-adjusted parameters.}
 \end{subfigure}
 \begin{subfigure}[b]{0.49\linewidth}
    \centering
    \includegraphics[width=0.9\columnwidth]{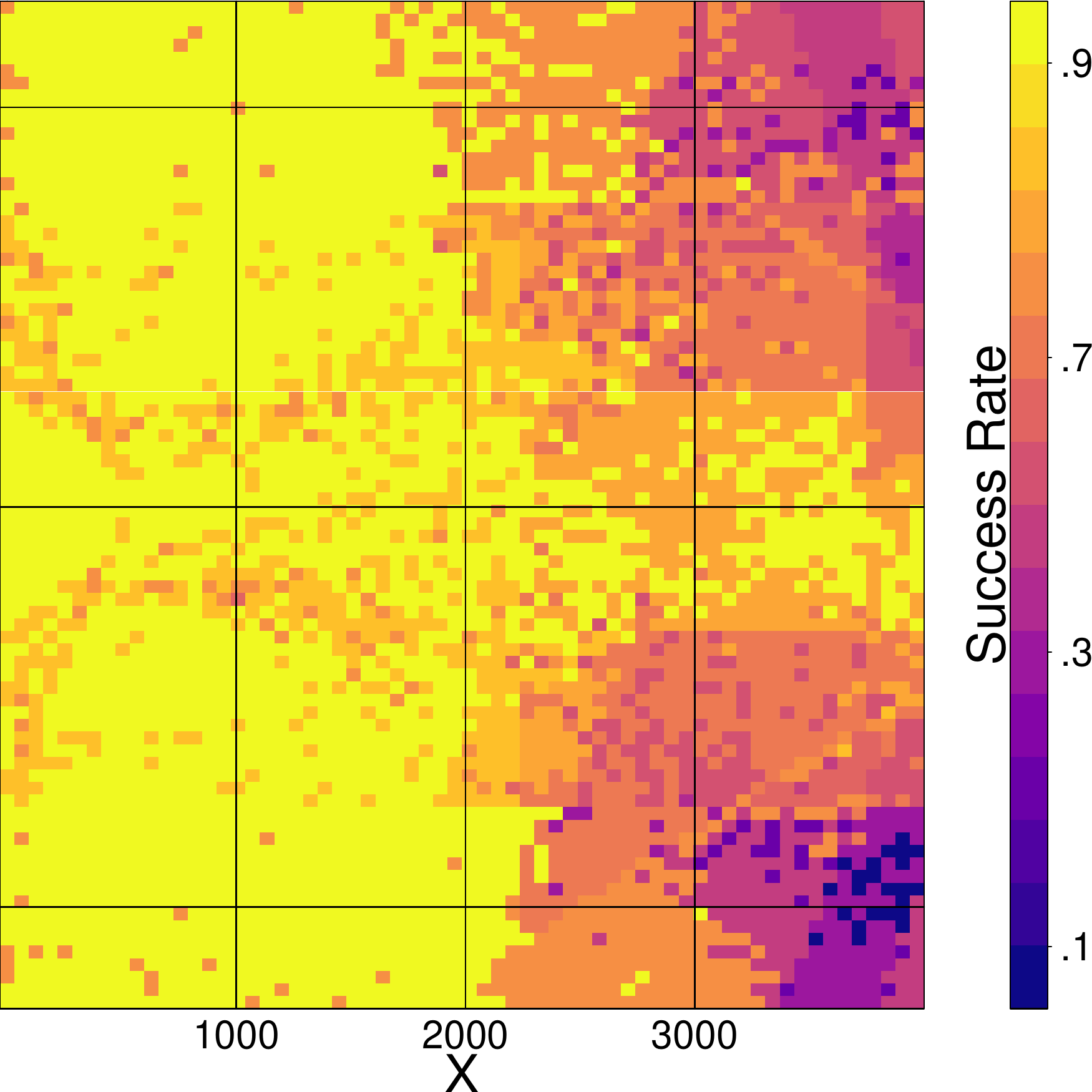}
    \caption{\technique{}-adjusted parameters.}
  \end{subfigure}
  \vspace{-.7em}
  \caption{Attacker success rate with respect to different initial ball
  positions. The corrections are marked with a cross.}
  \figlabel{heatmap}
\end{figure}

\begin{table}
  \centering
 \begin{tabular}{| l | r | r | r | r | r | r |}
  \hline
  \multirow{ 2}{*}{\textbf{RSM}} &
  \multirow{ 2}{*}{\textbf{Params}} &
  \multicolumn{1}{l|}{\textbf{\technique{}}}  &
  \multicolumn{1}{l|}{\multirow{ 2}{*}{\textbf{Tests}}}&
  \multicolumn{3}{c|}{\textbf{Success Rates ($\%$)}} \\
  \cline{5-7}
  & & \textbf{Corrections} & & \textbf{Baseline} & \textbf{Expert} &
\textbf{\technique} \\
  \hline
  Attacker    & 12 & 2 & 57,600 & 42 & 44 & 89 \\
  \hline
  Deflector  & 5 & 3 & 16,776 & 1 & 65 & 80 \\
  \hline
  Docker     & 9 & 3 & 5,000 & 0 & 0 & 100 \\
  \hline
  Passing    & 5 & 2 & 17296 & 50.4 & 71.4 & 86.1 \\
  \hline
  \end{tabular}
  \vspace{-.7em}
  \caption{
  Success rates for baseline, expert, and \technique{} parameters.}
  \tablabel{RSM_stats}
  \vspace{-.7em}
\end{table}

The second major correction method used by \technique{} is \continue{s}. To evaluate
\continue{s} we first demonstrate the failings of premature \imm{s},
evaluate the performance change of \continue{s} alone, and finally show that
\continue{} performance coupled with generalization performance yields substantially
improved success rates over premature \imm{s}.

We evaluate the need for \continue{s} and their performance using the attacker,
the experimental procedure
described in \secref{controller_repair}, an initial parameter configuration which
results in premature transitions, and four sets of corrections:
\begin{inparaenum}[1)]
  \item A single negative constraint which negates the transition into kick,
  \item $26$ negative constraints and $1$ positive constraint generated using
    a \continue{},
  \item the \continue{} constraints $+20$ positive constraints representing
    pre-repair behavior that we want to maintain, and
  \item the \continue{} constraints, restricted to modifying a subclause
   of the kick transition by user specification.
\end{inparaenum}

The initial parameter set yields the heatmap shown in \figref{premature_kick},
which has degraded performance from nominal in several regions of the field yielding a
success rate of ~$78.5\%$.
To illustrate the need for \continue{s} we show the
change in success rate after a single negative correction in \figref{neg_cor}.
The change in success rate is minimal, comparable to the magnitude of the noise in the system, yielding a success rate of ~$78.6\%$ over all trials,
and further repair would require manual iteration. In comparison, \figref{cont_cor}
shows the results of the single \continue{}, which shows a more
substantial change in performance overall,
but some performance degradation resulting from poor generalization, with a final success rate of ~$78.3\%$.

\subsection{Continue Corrections}
\begin{figure}
  \begin{adjustwidth}{-2.5cm}{-2.5cm}
  \centering
 \begin{subfigure}[b]{0.3\linewidth}
    \centering
    \includegraphics[width=0.9\columnwidth]{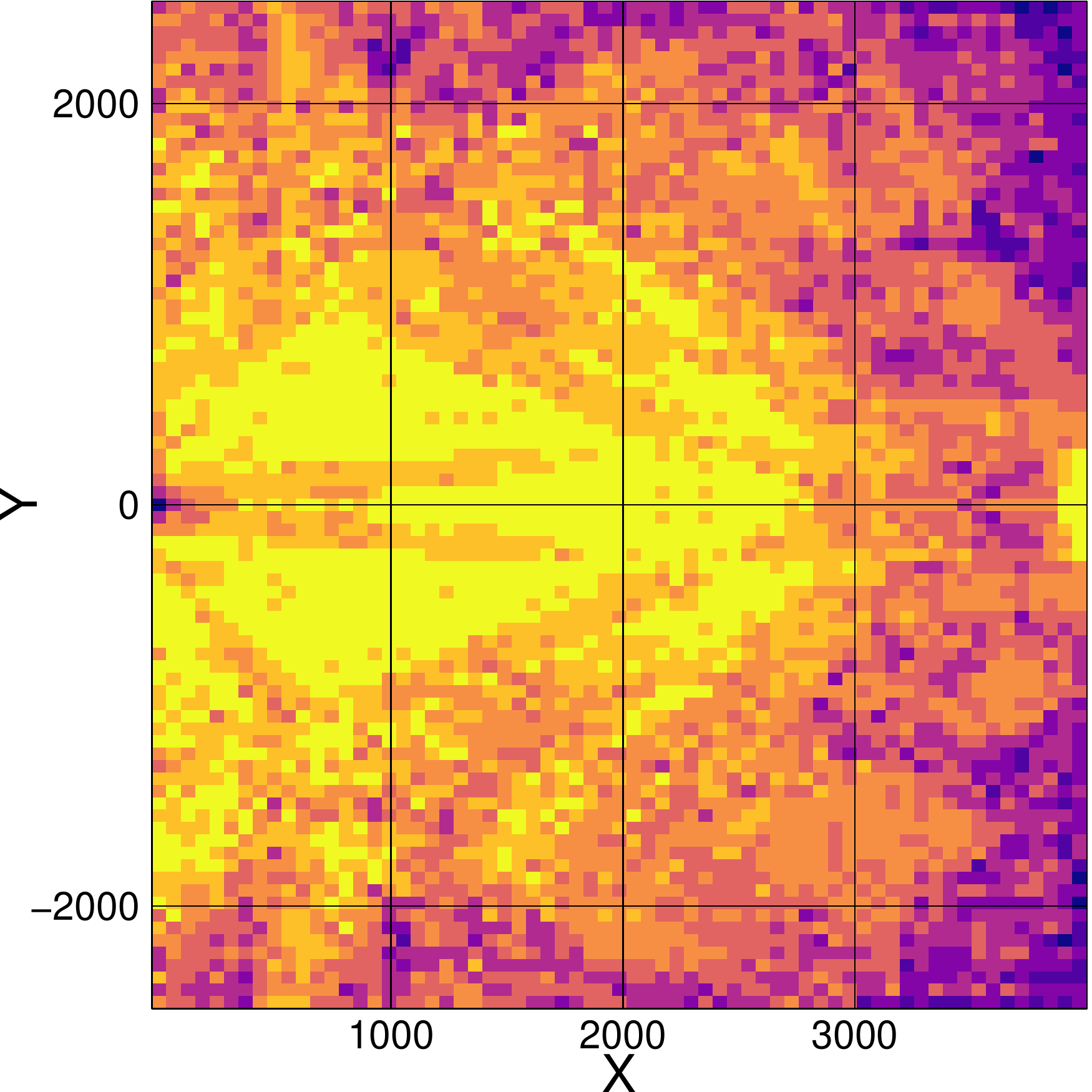}
    \caption{Premature kicking}
    \figlabel{premature_kick}
 \end{subfigure}
 \begin{subfigure}[b]{0.3\linewidth}
    \centering
    \includegraphics[width=0.9\columnwidth]{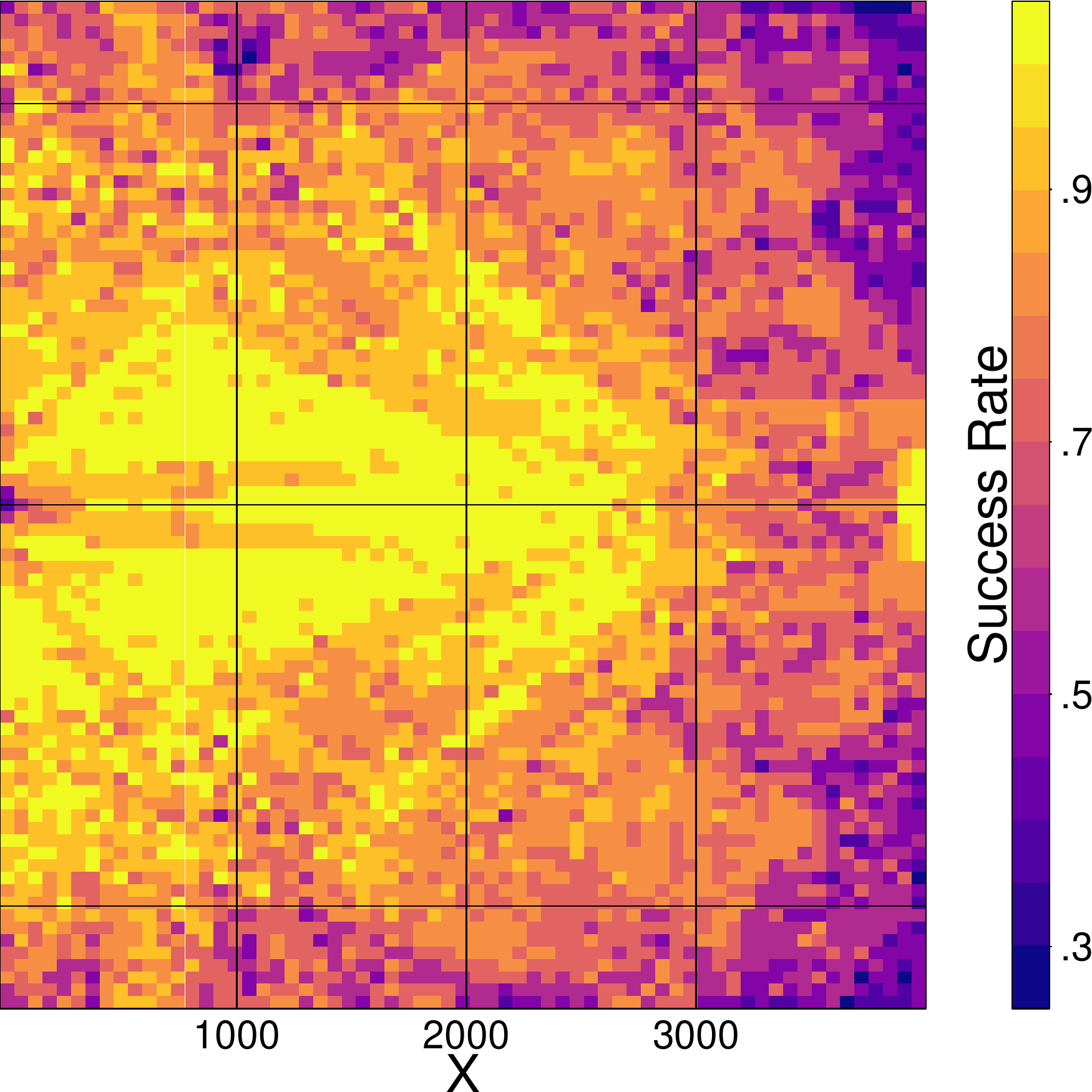}
    \caption{Single "No Kick" Correction}
    \figlabel{neg_cor}
  \end{subfigure}
  \begin{subfigure}[b]{0.3\linewidth}
    \centering
    \includegraphics[width=0.9\columnwidth]{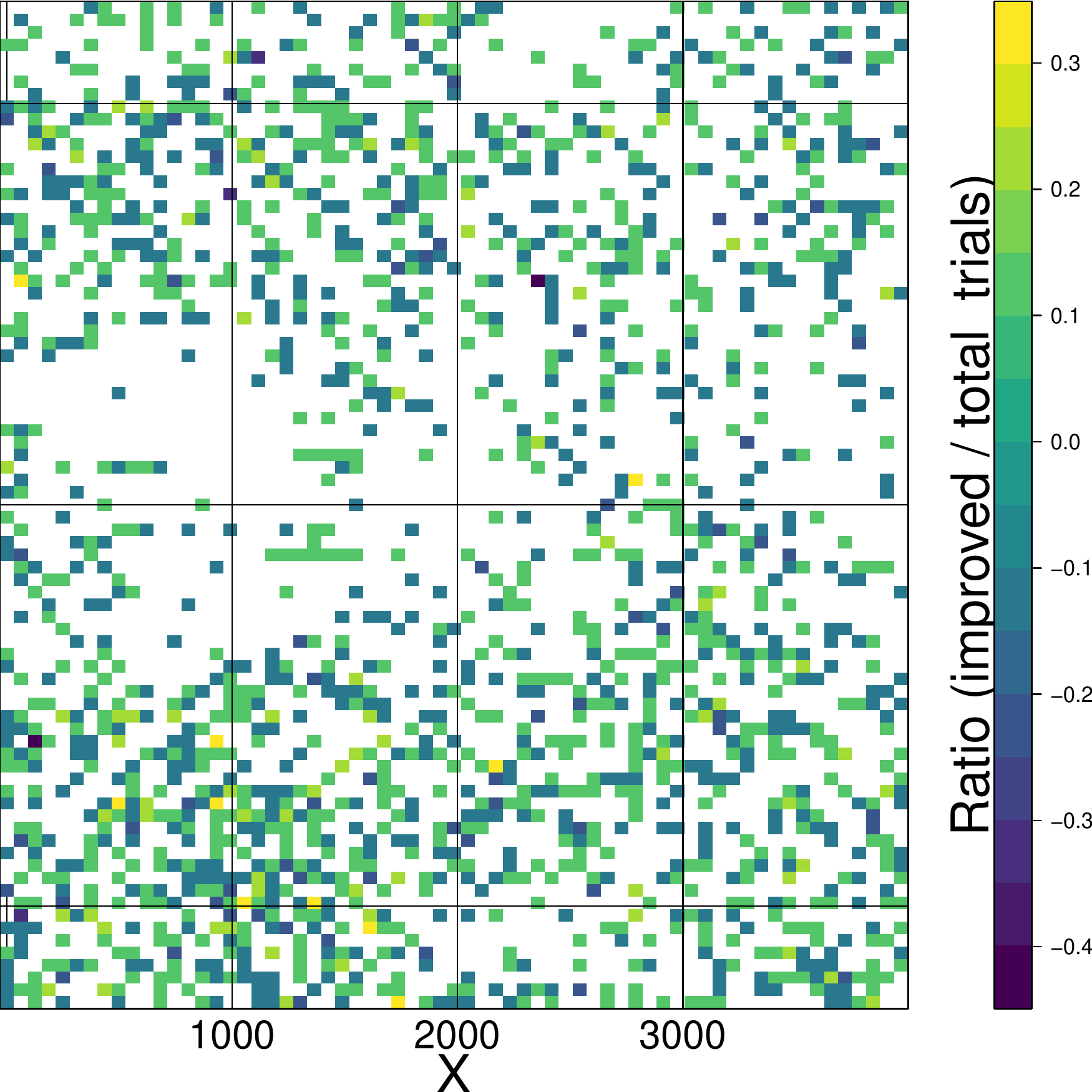}
    \caption{Change in success rate for "no kick".}
    \figlabel{neg_cor_change4}
  \end{subfigure}
  \vspace{-.7em}
  \centering
  \caption{Success rate for the attacker with an imperfect configuration and
  after one negative correction, as well as change in success rate between the two. White
  space represents locations with no change in success rate.}
  \end{adjustwidth}
\end{figure}

We evaluated two methods for improving generalizability, user specification
of clauses to adjust, and additional kick constraints. To test user specification
we took the \continue{} from \figref{neg_cor} and restricted the clause
adjusted to line~\ref{line:yLoc} of \figref{example-function}, which
contains one of the two parameters adjusted to yield these failures. For additional
kick constraints we sampled $20$ successful kick transitions at random from the
premature kicking scenario and used them alongside our \continue{}
for repairs. We show the results
for constrained clause adjustment and
additional kick constraints in \figref{clauseConst} and \figref{addKicks}
respectively. Both show much improved generalization over a single
continue correction, and improved success rate overall with a success rate of
$82.4\%$ for user guided clause adjustment, and $82.6\%$ with additional kick constraints.

The results of this evaluation show the failings of \imm{s}
when desired transition states cannot be found in a trace, and the ability
of \continue{s} to remedy this issue. While \continue{s} alone are overly
restrictive of the transitions being corrected, when coupled with techniques
that improve generlization \continue{s} yield improved parameter performance
while removing the need for arduous repeated experimentation.

\begin{figure}
  \begin{adjustwidth}{-1.5cm}{-1.5cm}
 \centering
  \begin{subfigure}[b]{0.3\linewidth}
    \centering
    \includegraphics[width=0.9\columnwidth]{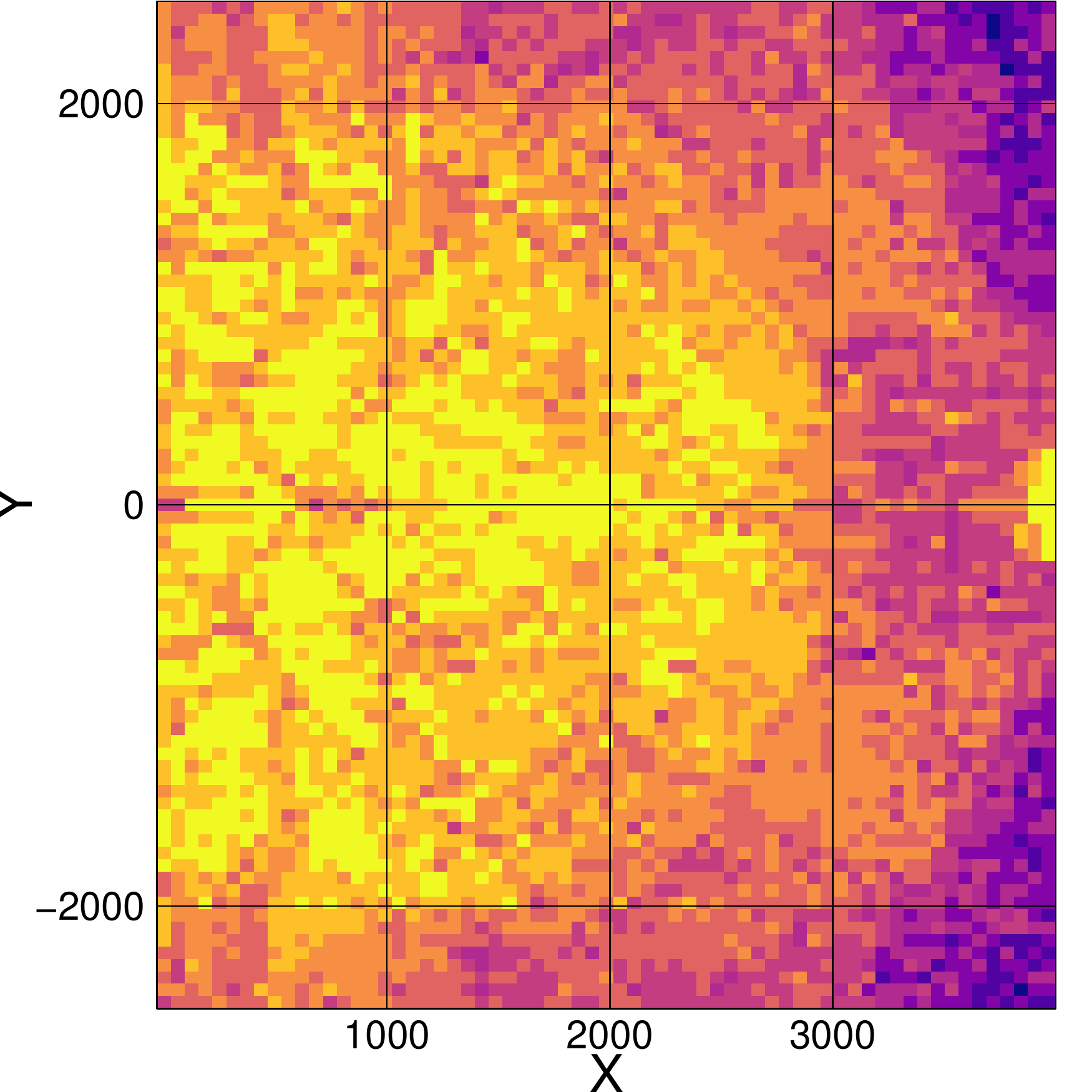} \\
    \includegraphics[width=0.9\columnwidth]{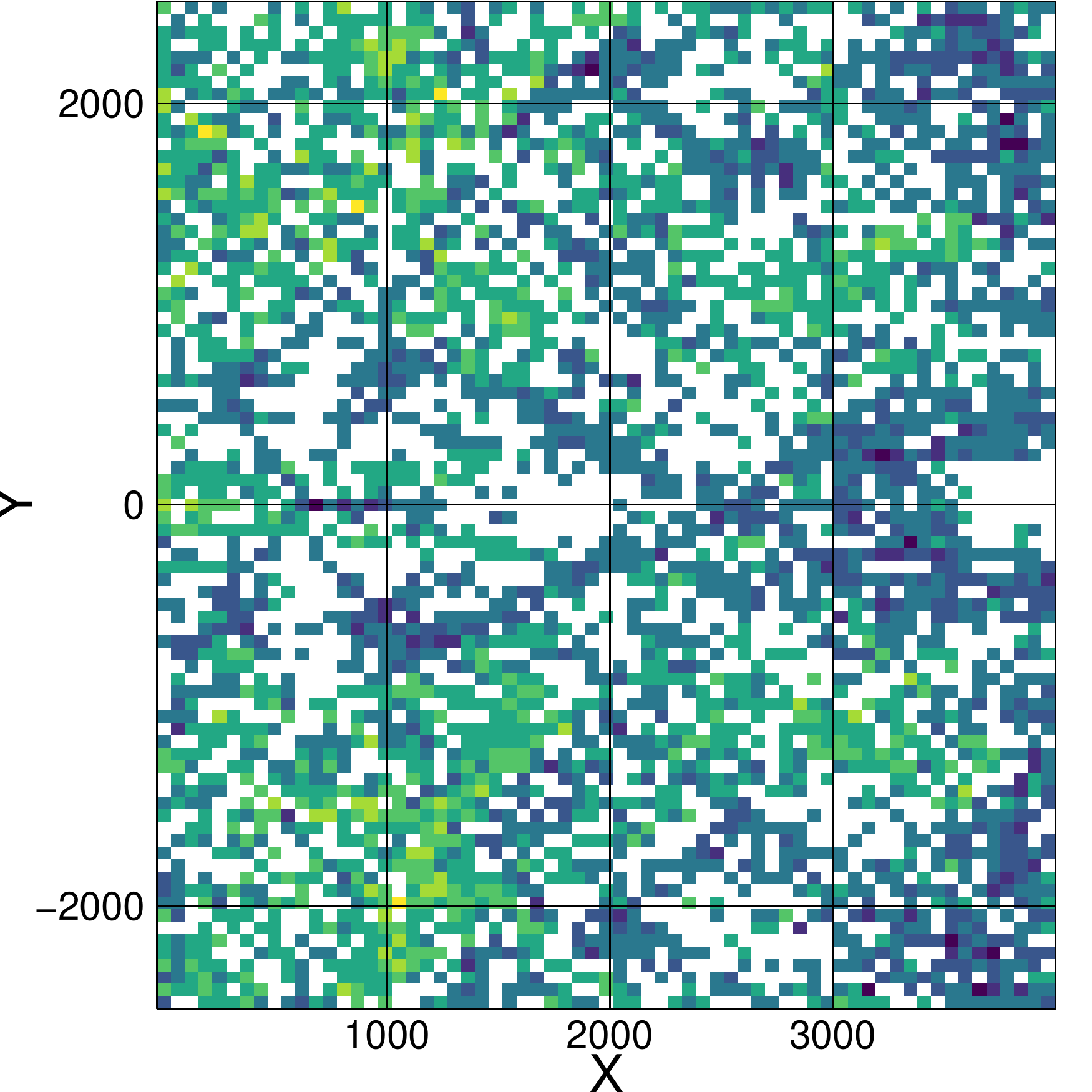}
    \caption{Continue Correction}
    \figlabel{cont_cor}
  \end{subfigure}
  \begin{subfigure}[b]{0.3\linewidth}
    \centering
    \includegraphics[width=0.9\columnwidth]{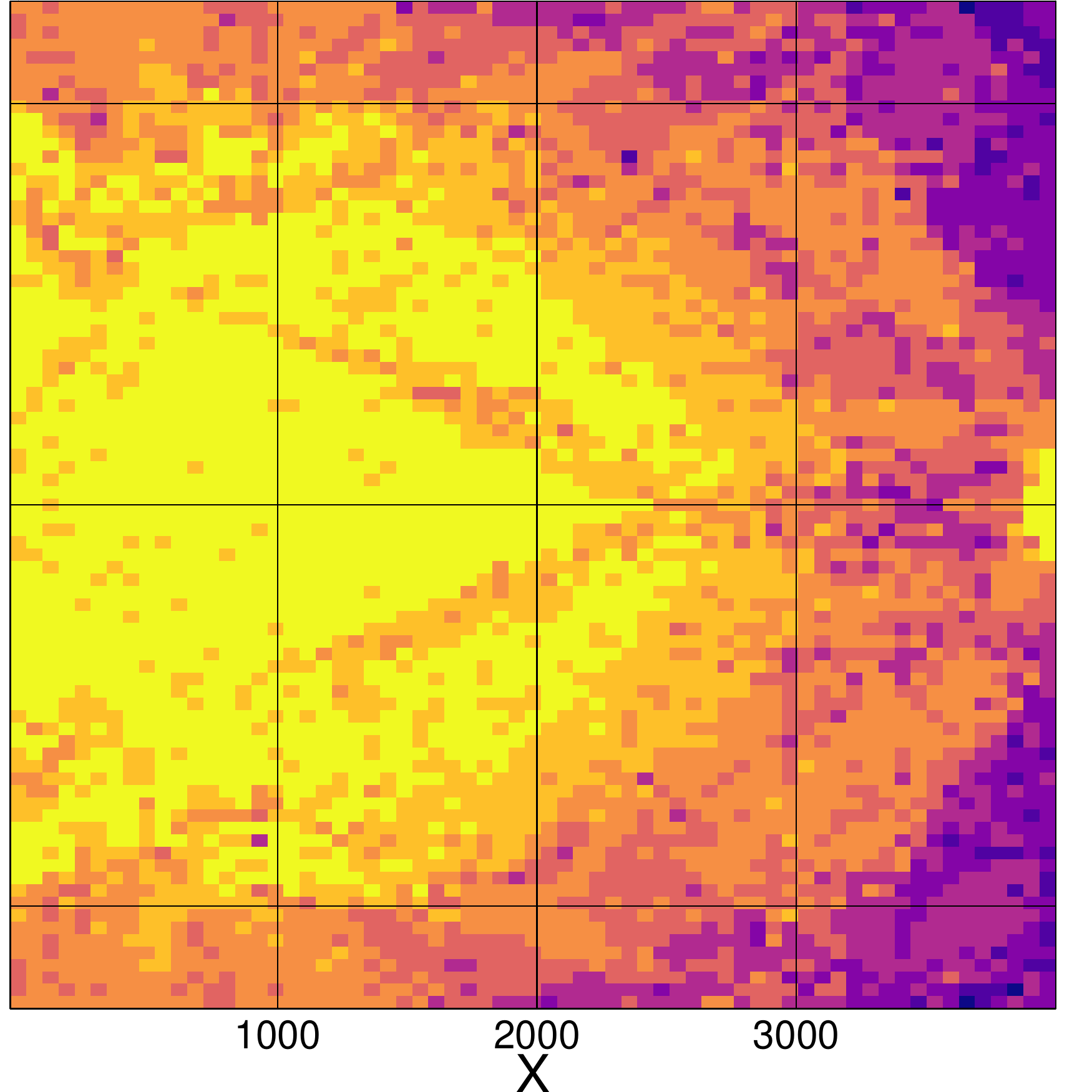} \\
    \includegraphics[width=0.9\columnwidth]{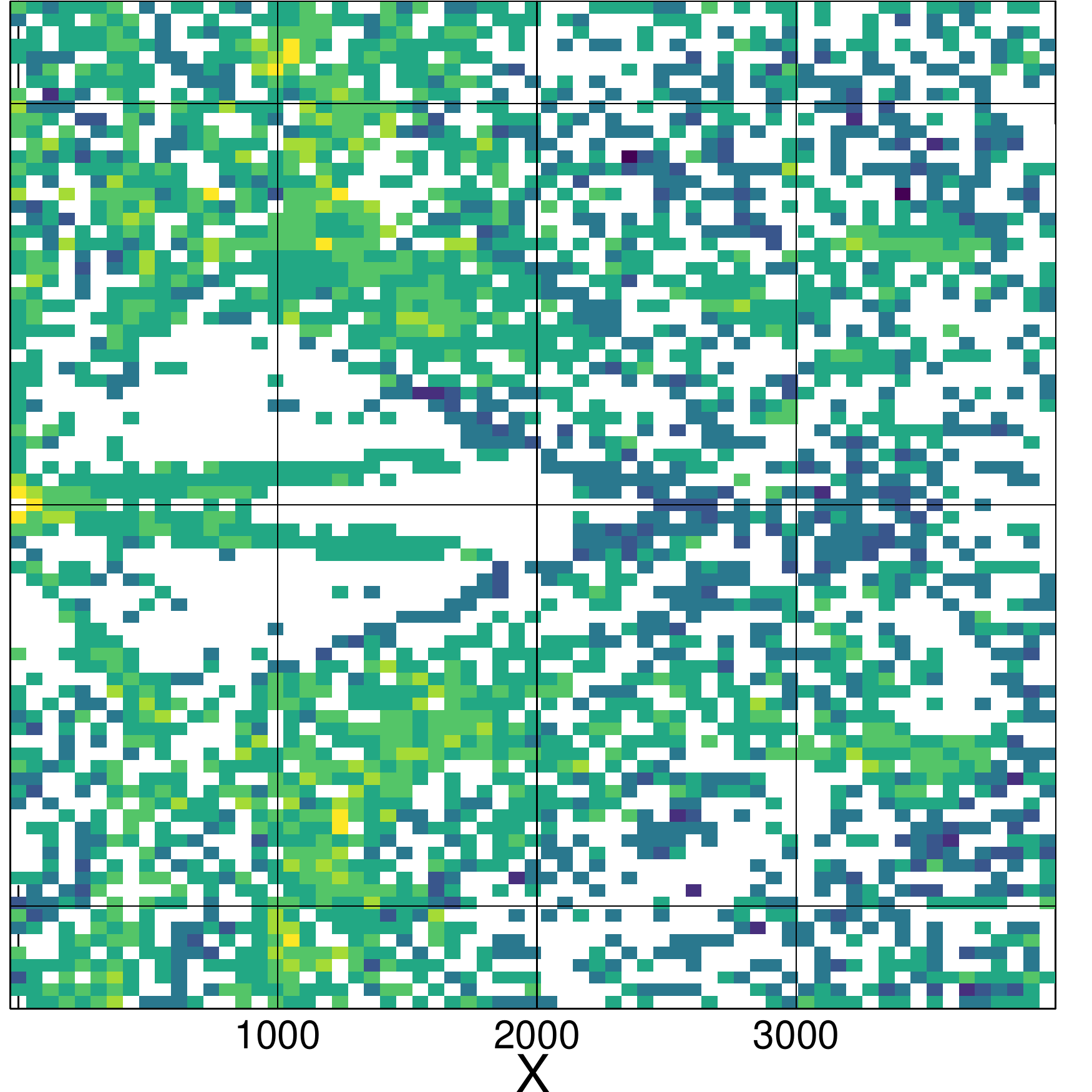}
    \caption{User Guided Adjustment}
    \figlabel{clauseConst}
  \end{subfigure}
  \begin{subfigure}[b]{0.3\linewidth}
    \centering
    \includegraphics[width=0.9\columnwidth]{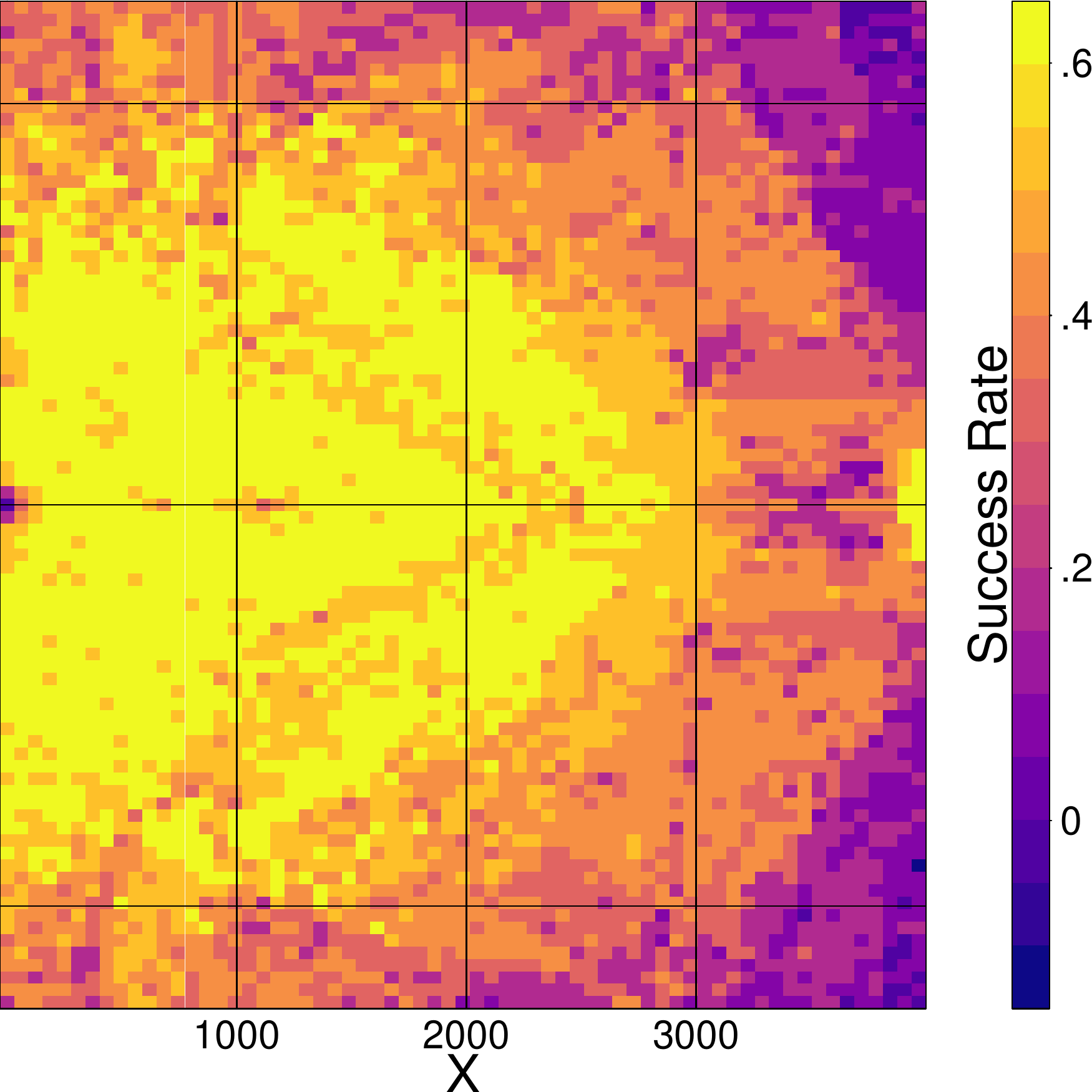} \\
    \includegraphics[width=.9\columnwidth]{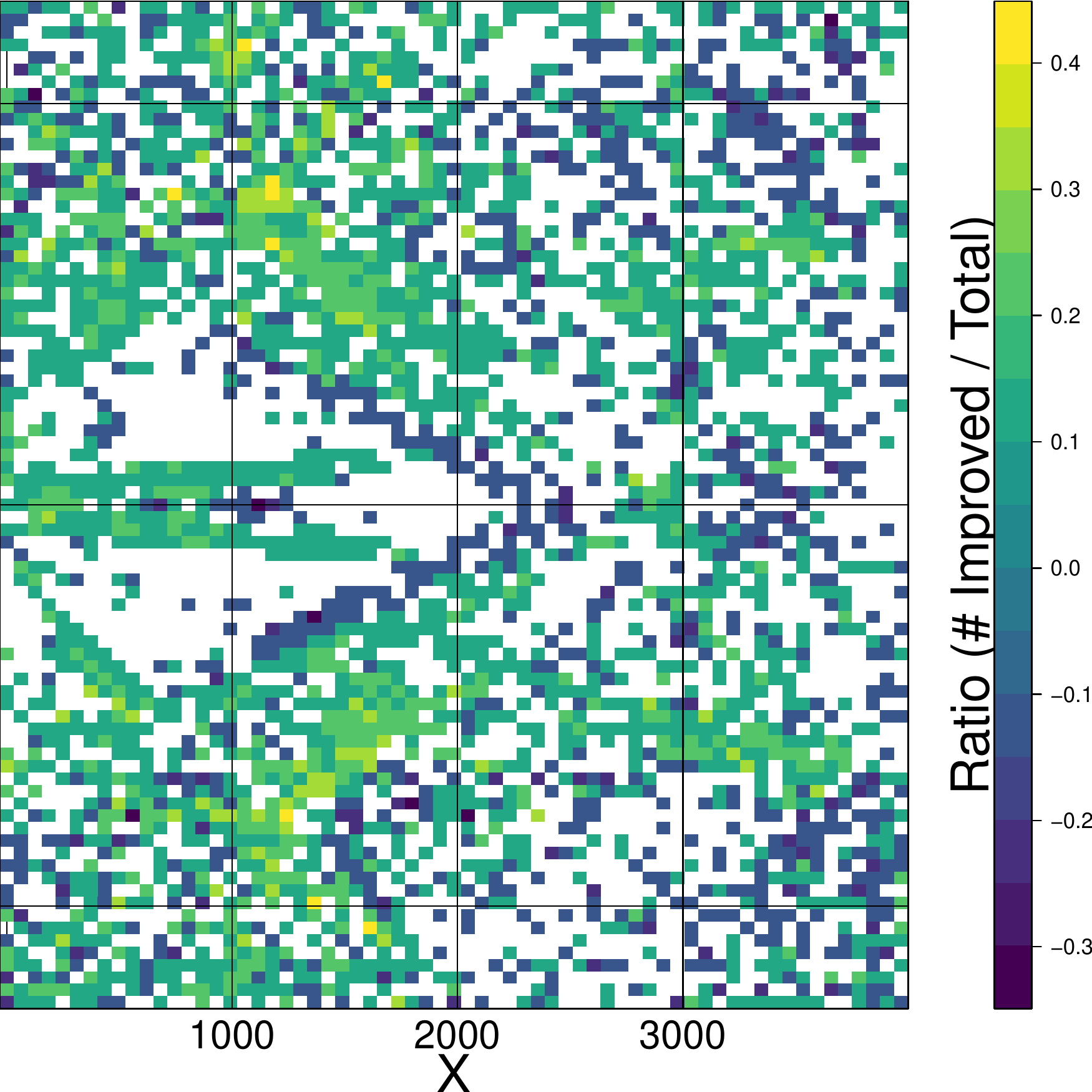}
    \caption{Additional kick constraints.}
    \figlabel{addKicks}
  \end{subfigure}
  \vspace{-.7em}
  \centering
  \captionsetup{justification=centering}
  \caption{Success rate for the attacker after repairs. The top row shows the full success rate
  heatmaps, and the bottom row visualizes the change in success rate compared to the premature kick scenario.}
  \figlabel{no_heatmaps}
  \end{adjustwidth}
\end{figure}

\subsection{Solution Exploration}
\seclabel{modelEnumResults}
\technique{} utilizes a variant on model enumeration in order to present various
possible solutions to a user. In order to evaluate the significance
of enumerated models versus the first \maxsmt{} solution we
use the attacker RSM and evaluate performance using varying ball positions
as described in \secref{controller_repair}. We start with a parameter set
that yields a roughly 30\% success rate, and a set of ten corrections with
several conflicts.

We show the success rate for the initial
configuration in \figref{enum_degraded}, and the success rate for three different
enumerated models in \figref{enum_models}. The initial configuration performs very
poorly in most cases, and all three enumerated models represent a significant
improvement over the baseline. \figref{enum_0} and \figref{enum_2} lead to very similar
performance, as the final solutions are not substantially different,
both yielding a success rate of roughly $85$\%. However, the
\figref{enum_1} has different performance characteristics in different regions of
the field, particularly near the sides of the goal,
but a lower success rate overall
of $80$\%. While the
solution found in this third iteration is not the optimal solution
in terms of cost or success rate, it represents a reasonable alternative solution when
user has supplied conflicting corrections without comparative weights, which
favors success near the goal over other locations.

The four heatmaps in \figref{enum_models},
and their corresponding models shown in \tabref{model_deltas},
demonstrate the ability of \technique{} to present qualitatively and
quantitatively different models given a set of conflicting corrections without
specific user input on the relative importance of the corrections. All of
the models produced by \technique{} represent a significant performance increase
over the degraded parameter configuration, and the variation between them
represent different interpretations of the correction intent.


\begin{table}
  \centering
  \begin{tabular}{ | l | c | c | c | c | c | c | c | c | }
  \hline
  \textbf{Iteration} & \textbf{$\delta^0$} & \textbf{$\delta^1$} & \textbf{$\delta^2$} &
  \textbf{$\delta^3$} & \textbf{$\delta^4$} & \textbf{$\delta^5$} &  \textbf{$\delta^6$} &
  \textbf{$\delta^7$} \\
  \hline
  Degraded & 0 & 0 & 0 & 0 & 0 & 0 & 0 & 0 \\
  \hline
  1 & 12.861 & 4.445 & 13.891 & 464.106 & 0 & -500.05 & 40.38 & 0 \\
  \hline
  2 & 5.189 & -0.027 & 12.41 & 463.63 & 0 & 0 & 39.84 & 0 \\
  \hline
  3 & 12.444 & 3.029 & 12.476 & 463.689 & 0 & -500.05 & 39.0007 & 0 \\
  \hline
  \end{tabular}
  \vspace{-.5em}
  \caption{$\delta^i$ values for enumerated models results shown in \figref{enum_models}.}
  \tablabel{model_deltas}
  \vspace{-1em}
\end{table}

\begin{figure}
  \begin{adjustwidth}{-3.5cm}{-3.5cm}
 \centering
   \begin{subfigure}[b]{0.2\linewidth}
    \centering
    \includegraphics[width=0.9\columnwidth]{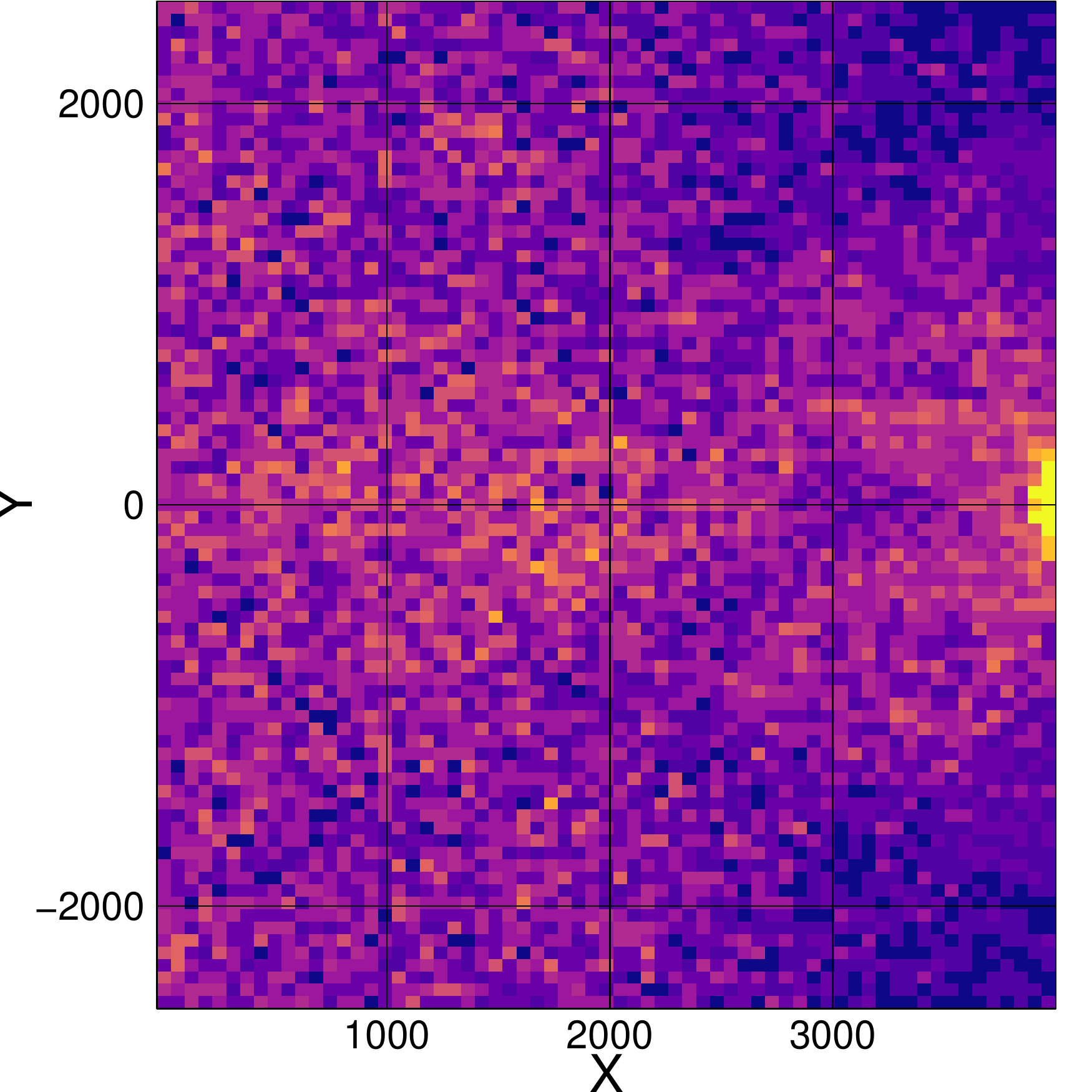}
    \caption{Degraded performance.}
    \figlabel{enum_degraded}
  \end{subfigure}
  \begin{subfigure}[b]{0.2\linewidth}
    \centering
    \includegraphics[width=0.9\columnwidth]{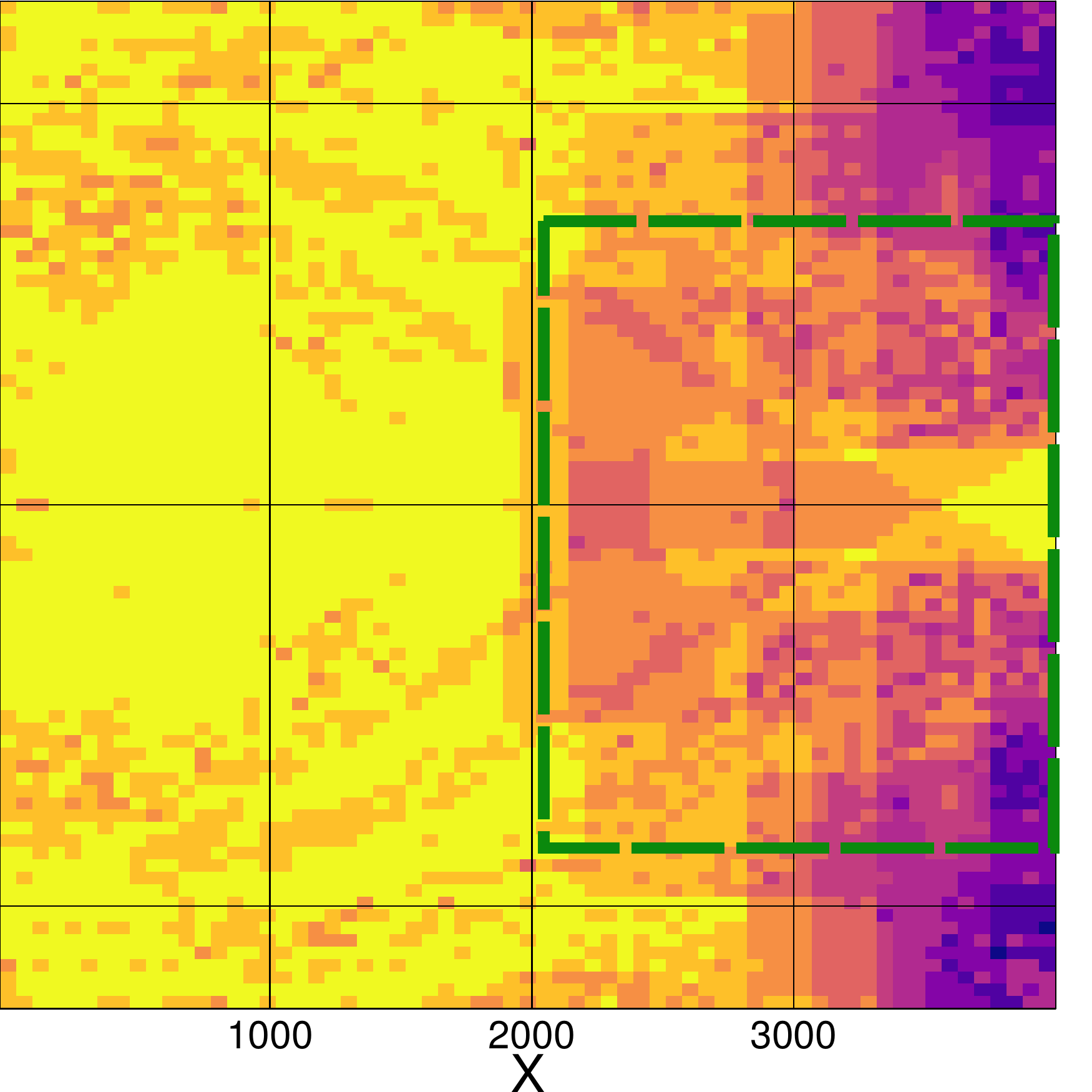}
    \caption{First enumerated model.}
    \figlabel{enum_0}
  \end{subfigure}
  \begin{subfigure}[b]{0.2\linewidth}
    \centering
    \includegraphics[width=0.9\columnwidth]{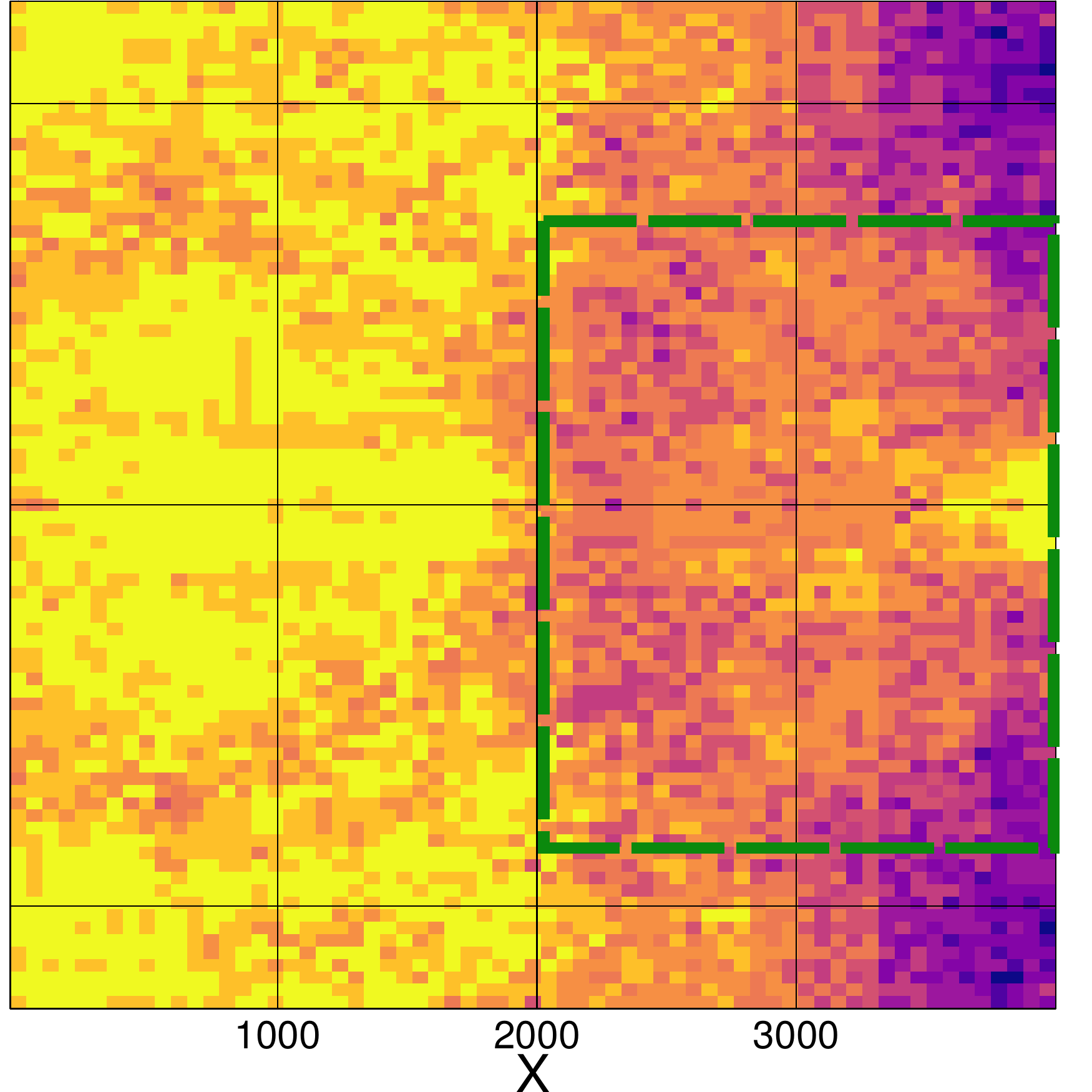}
    \caption{Second enumerated model.}
    \figlabel{enum_1}
  \end{subfigure}
  \begin{subfigure}[b]{0.2\linewidth}
    \centering
    \includegraphics[width=0.9\columnwidth]{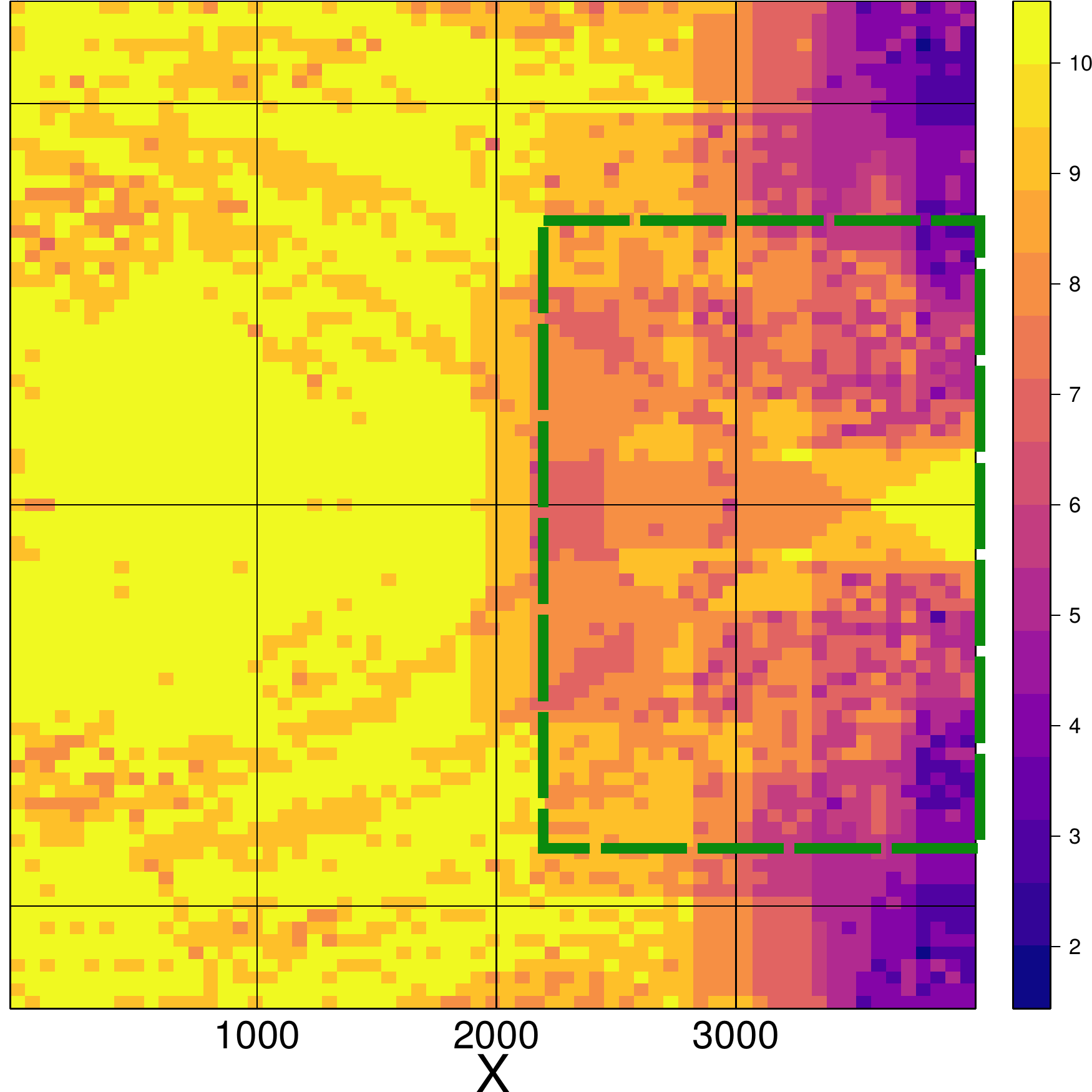}
    \caption{Third enumerated model.}
    \figlabel{enum_2}
  \end{subfigure}
  \vspace{-.7em}
  \centering
  \caption{Success rate heatmaps for degraded performance and three different
            solutions found via solution exploration. The green box highlights
            the region where performance changes between solutions.}
  \figlabel{enum_models}
  \end{adjustwidth}
\end{figure}

\subsection{Case Study: \technique{} In The Real World}
\seclabel{real_world_repair}

To evaluate \technique{} in the real world, we follow the same procedure that
experts use (summarized in \tabref{real_tuning}): we develop the Attacker in a simulator, we adjust parameters
until it performs well in simulation, and then we find
that it performs poorly in the real world.
To evaluate the success rate of the attacker in the real world,
we start the ball from 18 positions on the field and repeat each position five times with the same velocity (\ie{} 90 trials).
The parameters from
simulation have a 25\% success rate. Using the
execution logs of this experiment, we apply
\technique{} with three corrections. The adjusted parameters increase the success rate
to 73\%. In practice, an expert would iteratively adjust parameters, so we
apply \technique{} again with 2 more corrections, which increases the success rate to 86\%.
Finally, our group has an attacker that we tested and optimized
extensively for RoboCup 2017, where it was part of a team that won the lower bracket.
This \emph{Competition Attacker} has additional states to handle special cases that do not occur in our tests. On our tests, the
Competition Attacker's success rate is 76\%. Therefore, with two iterations of
\technique{}, the simpler Attacker actually outperforms the Competition Attacker
in typical, real-world scenarios.

\begin{table}
  \centering
  \begin{tabular}{ | l | r |}
  \hline
  \textbf{Trial} & \textbf{Success Rate ($\%$)} \\
  \hline
  \cellcolor{blue!25} Competition Attacker & \cellcolor{blue!25} 75 \\
  \hline
  Parameters from Simulation & 24 \\
  \hline
  Real World \technique{} Tuning 1 & 73 \\
  \hline
  Real World \technique{} Tuning 2 & 85 \\
  \hline
  \end{tabular}
  \vspace{-.5em}
  \caption{Attacker success rates on a real robot.}
  \tablabel{real_tuning}
  \vspace{-1em}
\end{table}

\section{Conclusion}

In this article we presented a solver based repair technique for
Robot State Machines. Our method, SMT-based Robot Transition Repair
(\technique{}) is a semi-automatic white-box approach for adjusting the transition parameters in RSMs.
\technique{} leverages different types interaction methods for user provided
corrrections to handle different failure modes, and uses solution exploration
to generate repairs that best model the user intent.
We demonstrate that \technique{}:
\begin{inparaenum}[1)]
\item increases success rate for multiple behaviors;
\item finds new parameters quickly using a small number of annotations;
\item produces solutions which generalize well to novel situations; and
\item improves performance in a real world robot soccer application.
\end{inparaenum}
These results show the effectiveness logical solver based techniques
are applicable to real world problems in robotics.
Using transition function repair we showed that these techniques
are a viable approach for applications that are both conditional dependent
and require real-valued arithmetic optimization. Future work will seek to leverage solver
based repair and human in the loop techniques for broader problems in robotics.=

\section*{Acknowledgments}

This work is supported in part by AFRL and DARPA agreement \#FA8750-16-2-0042, and by
NSF grants CCF-1717636 and IIS-1724101.


\bibliography{references.bib}

\end{document}